\documentclass[landscape, letterpaper, 10 pt, conference]{ieeeconf}
\makeatletter
\let\NAT@parse\undefined
\makeatother

\IEEEoverridecommandlockouts  
\usepackage[pdftex]{graphicx}
\usepackage{subfig}
\usepackage{tikz}
\usetikzlibrary{shapes,backgrounds}
\usepackage{psfrag}
\usepackage{array}
\usepackage{multirow}
\usepackage{url}
\usepackage{bm}
\usepackage{bibentry}
\usepackage{gensymb} 
\usepackage{amsmath,amssymb,amsfonts,bm}
\usepackage{mathtools}
\usepackage{booktabs}
\usepackage[bottom]{footmisc}   
\usepackage{etoolbox} 
\robustify{\subref}
\usepackage{algorithmic}
\usepackage{textcomp}
\def\BibTeX{{\rm B\kern-.05em{\sc i\kern-.025em b}\kern-.08em
    T\kern-.1667em\lower.7ex\hbox{E}\kern-.125emX}}
\usepackage{fontawesome}
\usepackage{relsize}
\usepackage[pdftex]{hyperref}  
\hypersetup{
	bookmarks=true,
	unicode=false,
	pdftoolbar=true,
	pdfmenubar=true,
	pdffitwindow=false,
	pdfstartview={FitH},
	pdftitle={LIBRE: The Multiple 3D LiDAR Dataset},
	pdfauthor={A. Carballo \emph{et~al.}}
}

\newsavebox{\arrangebox}
\newlength{\arrangeht}

\newdimen\midrulewidth
\DeclareRobustCommand*{\IEEEauthorrefmark}[1]{\raisebox{0pt}[0pt][0pt]{\textsuperscript{\footnotesize\ensuremath{\ifcase#1\or *\or \dagger\or \ddagger\or%
				\mathsection\or \mathparagraph\or \|\or **\or \dagger\dagger%
				\or \ddagger\ddagger \else\textsuperscript{\expandafter\romannumeral#1}\fi}}}}

\graphicspath{{figs-pdf/}}
\newcommand{\etal}[1][{~et~al.}]{\emph{#1}}

\newcommand{\datasetname}{LIBRE}
\newcommand{\numlidars}{10}
\newcommand{\isep}{\hspace{-3pt}\mathrel{\raisebox{1pt}{{.}{.}}\hspace{-3pt}}\nobreak}
	
\newcommand\T{\rule{0pt}{2.6ex}}        

\newcommand{\upperbound}[1]{\mbox{${#1}$}}
\newcommand{\lowerbound}[1]{\mbox{$\text{-}{#1}$}}

\newcommand{\upperbounddegplus}[1]{\mbox{$\text{+}{#1}\degree$}\hspace{-0.3em}}
\newcommand{\lowerbounddeg}[1]{\mbox{$\text{-}{#1}\degree$}}
\newcommand{\vfovbounds}[2]{\mbox{$\left[\upperbound{#1},\hspace{-0.2em}\lowerbound{#2}\right]$}}

\newcommand{\vfovboundsdegplus}[2]{\mbox{$\left[\upperbounddegplus{#1},\lowerbounddeg{#2}\right]$}}

\begin{document}
\title{{\datasetname}: The Multiple 3D LiDAR Dataset}
\author{Alexander Carballo$^{1,4}$ 
\and
Jacob Lambert$^{2,4}$ 
\and
Abraham Monrroy~Cano$^{2,4}$ 
\and
David Robert Wong$^{1,4}$ 
\and
Patiphon Narksri$^{2,4}$ 
\and
Yuki Kitsukawa$^{2,4}$ 
\and
Eijiro Takeuchi$^{2,4}$ 
\and
Shinpei Kato$^{3,4,5}$ 
\and
Kazuya Takeda$^{1,2,4}$ 
\thanks{$^1$Institute of Innovation for Future Society, Nagoya University, Furo-cho, Chikusa-ku, Nagoya 464-8601, Japan.}%
\thanks{$^2$Graduate School of Informatics, Nagoya University, Furo-cho, Chikusa-ku, Nagoya 464-8603, Japan.}%
\thanks{$^3$Graduate School of Information Science and Technology, University of Tokyo, 7-3-1 Hongo, Bunkyo-ku, Tokyo, 113-0033, Japan.}%
\thanks{$^4$TierIV Inc., Nagoya University Open Innovation Center, 1-3, Mei-eki 1-chome, Nakamura-Ward, Nagoya, 450-6610, Japan.}%
\thanks{$^5$The Autoware Foundation, 3-22-5, Hongo, Bunkyo-ku, Tokyo, 113-0033, Japan.
Email: \tt\scriptsize \underline{alexander@g.sp.m.is.nagoya-u.ac.jp}, jacob.lambert@g.sp.m.is.nagoya-u.ac.jp, abraham.monrroy@tier4.jp, david.wong@tier4.jp, narksri.patiphon@g.sp.m.is.nagoya-u.ac.jp, yuki.kitsukawa@tier4.jp, takeuchi@g.sp.m.is.nagoya-u.ac.jp, shinpei@is.s.u-tokyo.ac.jp, kazuya.takeda@nagoya-u.jp}}

\maketitle

\begin{abstract}
In this work, we present {\datasetname}: LiDAR Benchmarking and Reference, a first-of-its-kind dataset featuring {\numlidars} different LiDAR sensors, covering a range of manufacturers, models, and laser configurations. Data captured independently from each sensor includes three 
different environments and configurations: \emph{static targets}, where objects were placed at known distances and measured from a fixed position within a controlled environment; \emph{adverse weather}, where static obstacles were measured from a moving vehicle, captured in a weather chamber where LiDARs were exposed to different conditions (fog, rain, strong light); 
and finally, \emph{dynamic traffic}, where dynamic objects were captured from a vehicle driven on public urban roads, multiple times at different times of the day, and including supporting sensors such as cameras, infrared imaging, and odometry devices. {\datasetname} will contribute to the research community to (1) provide a means for a fair comparison of currently available LiDARs, and (2) facilitate the improvement of existing self-driving vehicles and robotics-related software, in terms of development and tuning of LiDAR-based perception algorithms.
\end{abstract}


\section{Introduction}
\label{s:intro}
\begin{figure}[!htb]
	\centering
	\includegraphics[width=0.4\textwidth]{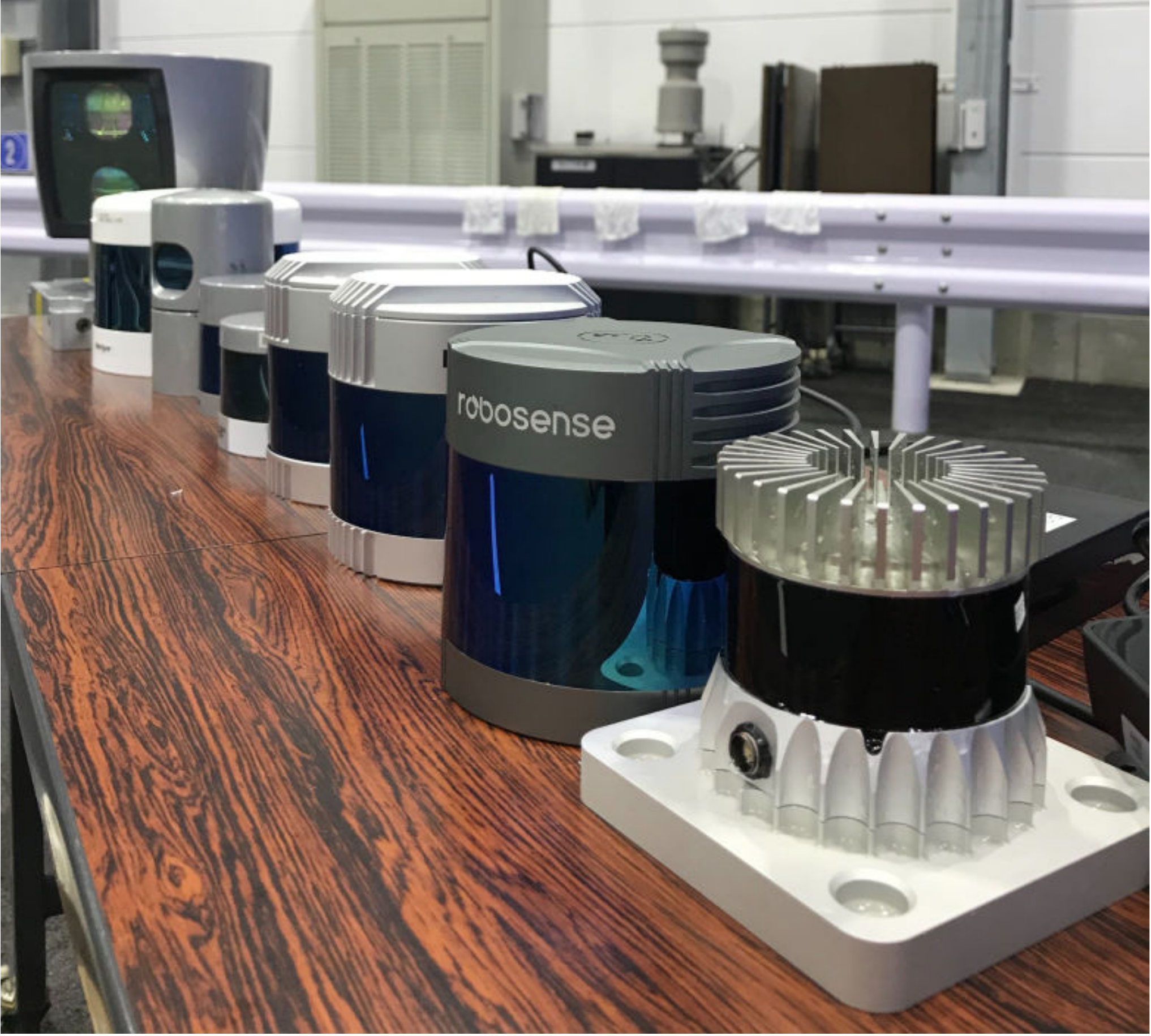}
	\caption{Multiple 3D LiDARs}
	\label{F:sensors}
	\vspace{-2em}
\end{figure}
LiDAR (\emph{Light Detection And Ranging}, sometimes \emph{Light Imaging Detection And Ranging} for the image-like resolution of modern 3D sensors) is one of the core perception technologies which has shaped the fields of Advanced Driver Assistance System (ADAS) and autonomous driving vehicles. While LiDARs are relative newcomers to the automotive industry when compared with radars and cameras, 2D and especially 3D LiDARs have demonstrated high measurement accuracy and illumination independent sensing capabilities for self-driving tasks\cite{thrun2006stanley}. Of course, not only used in automotive applications, LiDARs have been deployed in wheeled autonomous robots, drones, humanoid robots, consumer level applications, and at intersections in smart cities. The rapid development of research and industry relating to self-driving vehicles has created a large demand for such sensors. Depending on the individual perception application and operating domain, there are several key LiDAR performance attributes: measurement range, measurement accuracy, point density, scan speed and configurability, wavelength, robustness to environmental changes, form factor, and cost. As such, a large number of LiDAR manufacturers have emerged in recent years introducing new technologies to address such needs\cite{yole2018}.

With many different manufacturers and technologies becoming available, it is necessary to assess the perception characteristics of each device according to the intended application. In addition, while each LiDAR manufacturer subjects their products to quality tests (vibration and shock endurance, tolerance to electromagnetic interference (EMI), water and dust ingress protection (IP), operating temperature and pressure, measurement accuracy for different reflectors, etc.), LiDARs are meant for general use and not exclusively tested on vehicles. Furthermore, with LiDAR costs remaining high, it can be difficult to select the best LiDAR in terms of cost performance for a particular application.
{
	\begin{table*}[t]
		\begin{center}
			\footnotesize
			\setlength{\tabcolsep}{5pt}
			\begin{tabular}[c]{p{3.3cm}p{5.2cm}lllp{4.5cm}}
				\hline\hline
				Dataset & LiDAR(s) & Image & Labels & Diversity & Other sensors, notes\\
				\hline
				Stanford Track Collection\cite{stanford2011} & 1 (HDL-64S2) & - & 3D & E & GPS/IMU\\
				KITTI\cite{kitti2013} & 1 (HDL-64) & Yes & 2D/3D & E & 3x Cam (Stereo), GPS/IMU \\
				KAIST multispectral\cite{kaist2018} & 1 (HDL-32E) & Yes & 2D/3D & E/T & 2 Cameras, 1 Thermal (infrared) cam. IMU+GNSS \\
				nuScenes\cite{nuscenes2019} & 1 (HDL-32E) & Yes & 2D/3D & E/T/W & 6 Cameras, 5 RADARs, GNSS, IMU\\
				H3D\cite{h3d} & 1 (HDL-64S2) & Yes & 2D/3D & E & 3 Cameras, IMU+GNSS \\
				ApolloScape\cite{apolloscape2018} & 2 (2x Riegl VUX-1HA) & Yes & 2D/3D & E/T/W & Depth Images, GPS/IMU, \\
				LiVi-Set\cite{li-vi2018} & 2 (HDL-32E, VLP-16${}^{\bm{a}}$) & Yes & - & E & Dash-board camera, CAN (driving behavior dataset)\\
				ArgoVerse\cite{argoverse} & 2 (2x VLP-32C) & Yes & 2D/3D & E &  7 Cameras ring, 2 Stereo cams, GNSS\\
				FORD Campus\cite{ford2011} & 3 (HDL-64S2, 2x Riegl LMS-Q120) & Yes & - & E & Camera, $360\degree$ cam., IMU, INS\\
				Oxford RobotCar\cite{RobotCarDatasetIJRR} & 3 (2x SICK LMS-151 (2D), SICK LD-MRS (3D)) & Yes & - & E/T/W & 3x Cameras, Stereo cam., GPS \\
				Lyft\cite{lyft2019} & 3 (2 Pandar40 + 1 Pandar40 in Beta\_V0, and 2 Pandar40 + 1 Pandar64 in Beta\_Plus) & Yes & Yes & E & 6 Cameras, IMU, INS\\
				Waymo\cite{waymo2019} & 5 (1 $360\degree$ $75m$ range, 4x ``HoneyComb'' $20m$ range${}^{\bm{b}}$) & Yes & 2D/3D & E/T/W & 5 Cameras \\
				DENSE\cite{dense2019} & 2 (HDL-64S3, VLP-32C) & Yes & 2D/3D & E/T/W & Stereo Camera, Gated Camera, FIR Camera, Radar, laser illumination, weather station \\
				A2D2\cite{geyer2020a2d2} & 5 (5x VLP-16) & Yes & 2D/3D & E/W & 6x Cameras \\
				{\textbf{{\datasetname} (ours)}} & {\numlidars} (VLS-128\protect\footnotemark, HDL-64S2, HDL-32E, VLP-32C, VLP-16, Pandar64, Pandar40P, OS1-64, OS1-16, RS-Lidar32) & Yes & 2D/3D\protect\footnotemark & E/T/W & Camera, IMU, GNSS, CAN, $360\degree$ 4K cam., Event cam., Infrared cam., 3D pointcloud map, Vector map \\
				\hline
			\end{tabular}
			\caption{Publicly available datasets featuring LiDARs (arranged chronologically and by number of LiDARs, names of LiDAR manufacturers are omitted for those models in this study). Diversity refers to changes in the data collected, as in types of environments (E), times of day (T), weather conditions (W). ${}^{\bm{a}}$The authors in \cite{li-vi2018} state they only used the HDL-32E. ${}^{\bm{b}}$LiDARs proprietary and developed by Google/Waymo.}
			\label{tab:datasets}
			\vspace{-2em}
		\end{center}
	\end{table*}
}
In this study, we aim to collect data to enable the attribute analysis of several 3D LiDARs for applications in autonomous driving vehicles. We capture data to evaluate LiDARs in terms of: measurement range, accuracy, density, object detection, mapping and localization, and robustness to weather conditions and interference. During our study we collected a large dataset of vehicle-mounted LiDARs both in normal traffic scenes, and in a controlled chamber for testing performance in adverse weather conditions.
\addtocounter{footnote}{-1}
\footnotetext{In addition to the VLS-128, the Velodyne Alpha Prime will be also added to the dataset.}
\addtocounter{footnote}{1}
\footnotetext{At the time of writing, 2D/3D data labeling is ongoing. Labels will be included for a subsets of the dynamic traffic data.}

Following data capture in the above environments, we released the {\datasetname} dataset covering multiple 3D LiDARs.\footnote{A teaser of {\datasetname} dataset was released on January 28th, 2020 at \url{https://sites.google.com/g.sp.m.is.nagoya-u.ac.jp/libre-dataset}. The full set will be released during 2020. For additional details, please refer to the complementary video available at \url{https://youtu.be/rWyecoCtKcQ}.} It features {\numlidars} LiDARs, each one a different model from diverse manufacturers. Fig.~\ref{F:sensors} shows some of the 3D LiDARs used in our evaluations. The {\datasetname} dataset includes data from three  
different environments and configurations:
\begin{itemize}
	\item \emph{Dynamic traffic}: dynamic traffic objects (vehicles, pedestrians, bicycles, buildings, etc.) captured from a vehicle driving on public urban roads around Nagoya University
	\item \emph{Static targets}: static objects (reflective targets, black car and a mannequin), placed at known controlled distances, and measured from a fixed position
	\item \emph{Adverse weather}: static objects placed at a fix location and measured from a moving vehicle while exposing the LiDARs to adverse conditions (fog, rain, strong light)
\end{itemize}
The contributions of this work are summarized as follows. We introduce the {\datasetname} dataset including data from {\numlidars} different LiDARs in the above environments and configurations. We present a quantitative summary of performance of the different LiDARs in terms of range and density for static targets, and a qualitative evaluation of response to adverse weather conditions. While this paper offers some limited analysis of the large amount of data captured, the main contribution is the publishment of a novel and openly available dataset which will allow many researchers to perform more detailed analysis and comparisons.

This paper is structured as follows: Section~\ref{s:relworks} presents related datasets featuring LiDARs, while Section~\ref{s:dataset} describes our dataset. Section~\ref{s:env-dynamic} presents results on dynamic traffic scenes, Section~\ref{s:env-static} on static evaluations, and Section~\ref{s:env-weather} is about weather chamber tests. Finally, this paper is concluded in Section~\ref{s:concl}.
{
	\begin{table*}[t]
		\begin{center}
			\footnotesize
			\setlength{\tabcolsep}{4pt}
			\begin{tabular}{p{0.09302\textwidth}|p{0.05632\textwidth}p{0.08058\textwidth}p{0.084395\textwidth}p{0.05414\textwidth}p{0.04832\textwidth}|p{0.0532\textwidth}p{0.06395\textwidth}|p{0.07958\textwidth}p{0.08558\textwidth}|p{0.06395\textwidth}}
				\hline\hline \T
				& \multicolumn{5}{c|}{Velodyne} & \multicolumn{2}{c|}{Hesai} & \multicolumn{2}{c|}{Ouster} & RoboSense \\ \T
				&
				\begin{minipage}{0.05632\paperwidth}
					\raggedright
					\includegraphics[width=0.041746\paperwidth]{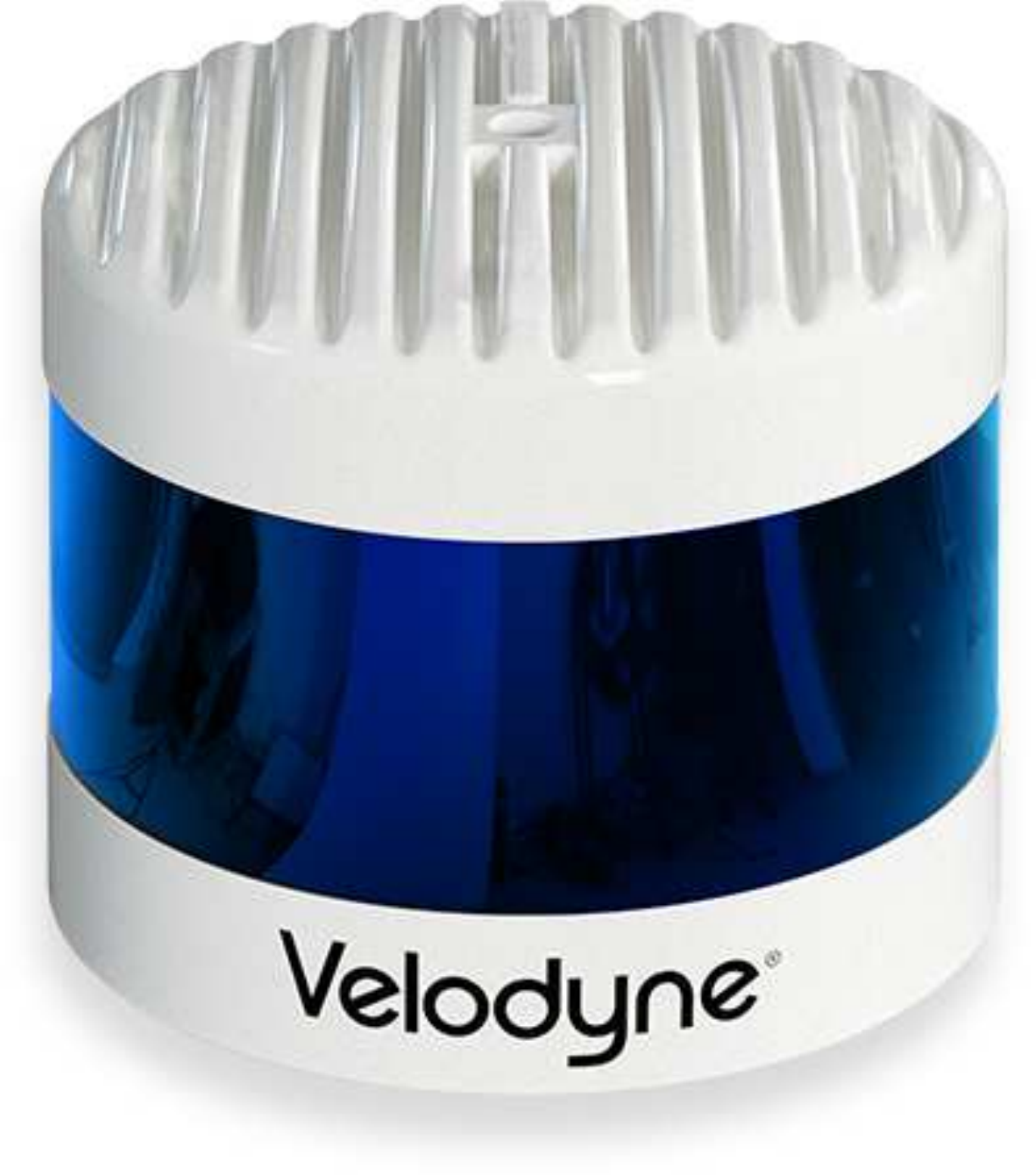}\\
					VLS-128$^{\bm{*}}$\\\cite{alphaprime}
				\end{minipage} & 
				\begin{minipage}{0.08058\paperwidth}
					\raggedright
					\includegraphics[width=0.056376\paperwidth]{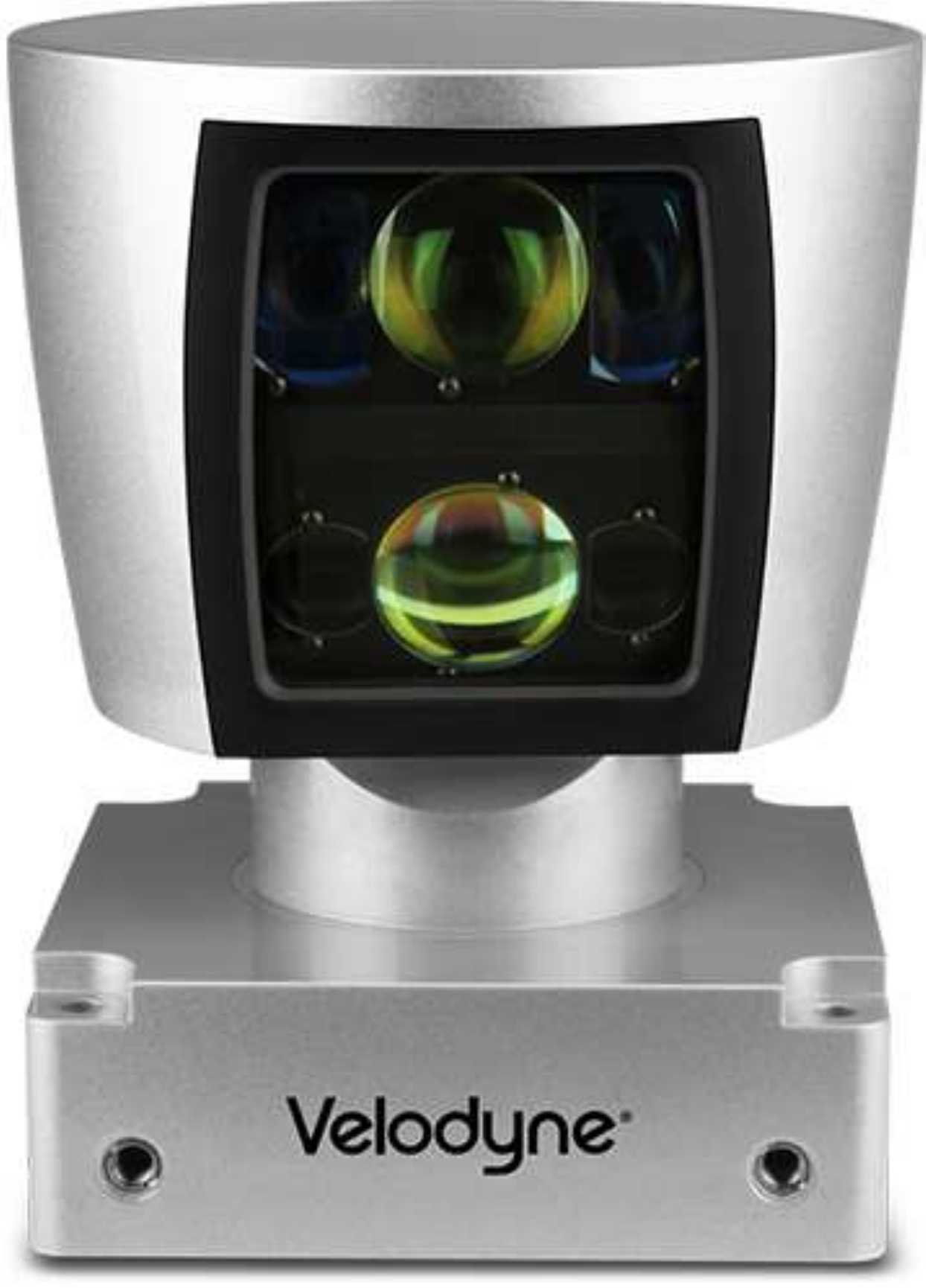}\\
					HDL-64S2\cite{hdl64s2}
				\end{minipage} &
				\begin{minipage}{0.081395\paperwidth}
					\raggedright
					\includegraphics[width=0.021516\paperwidth]{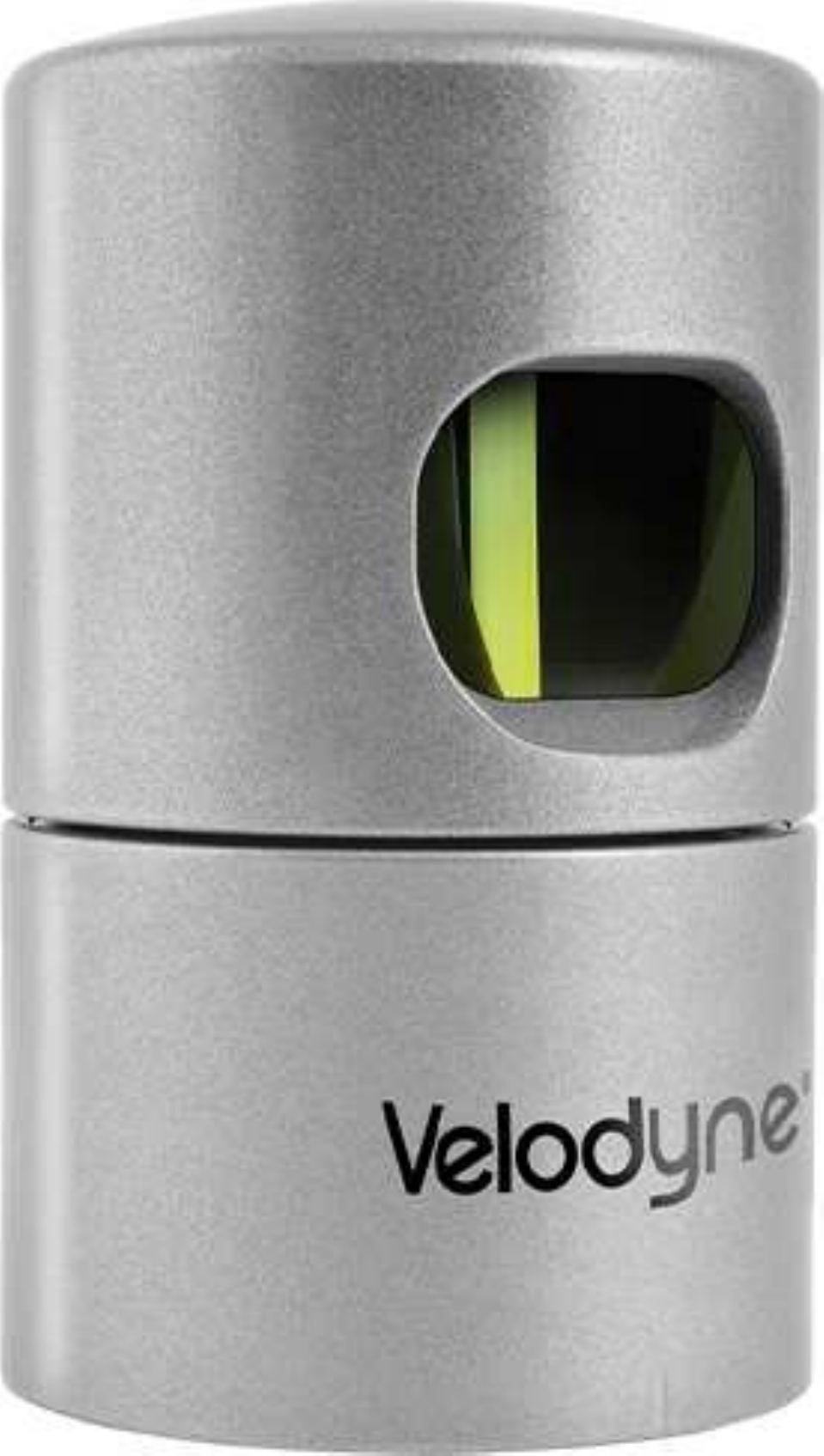}\\
					HDL-32E\\\cite{hdl32e}
				\end{minipage} &
				\begin{minipage}{0.06395\paperwidth}
					\raggedright
					\includegraphics[width=0.02598\paperwidth]{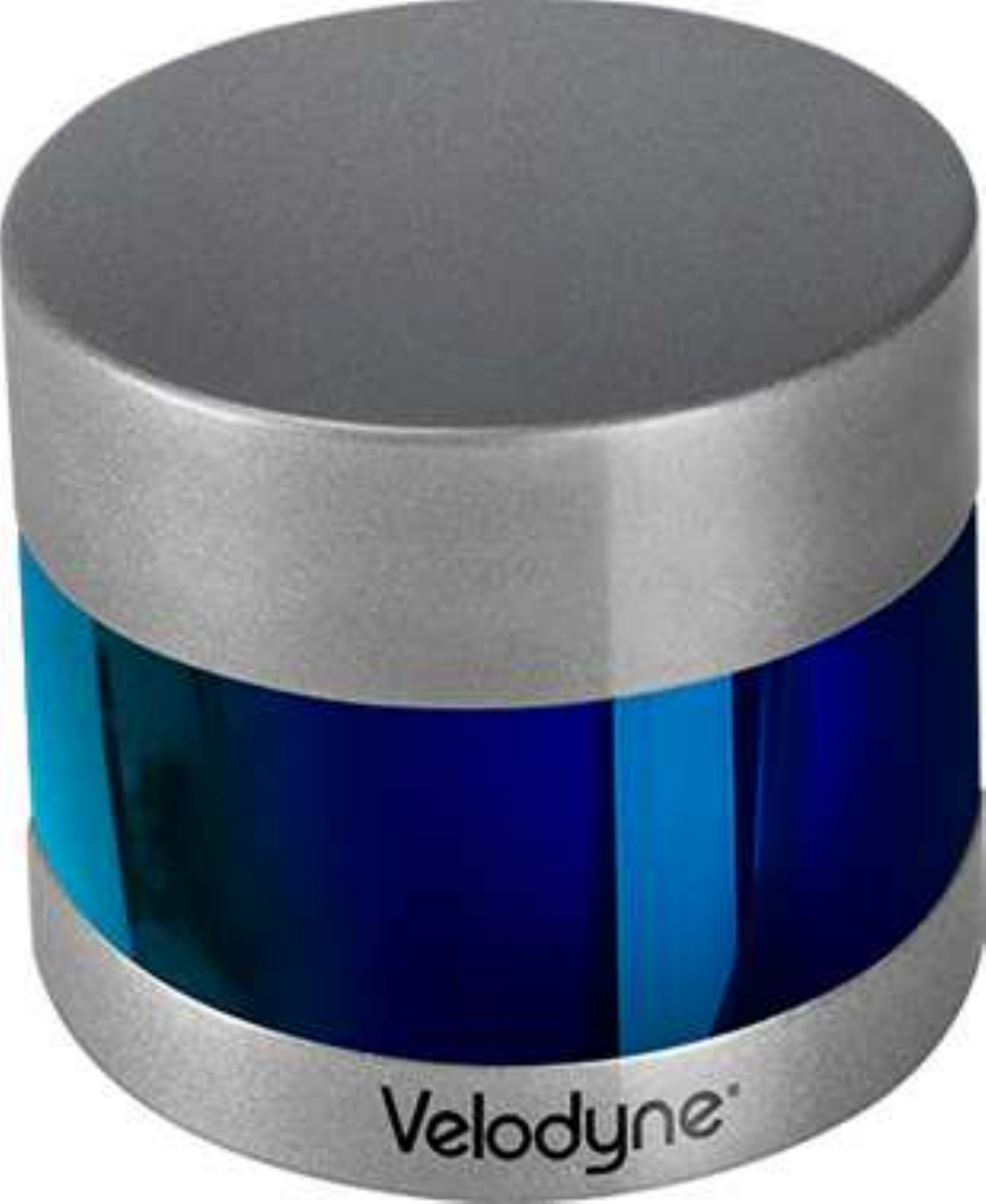}\\
					VLP-32C\\\cite{vlp32c}
				\end{minipage} &
				\begin{minipage}{0.05814\paperwidth}
					\raggedright
					\includegraphics[width=0.02606\paperwidth]{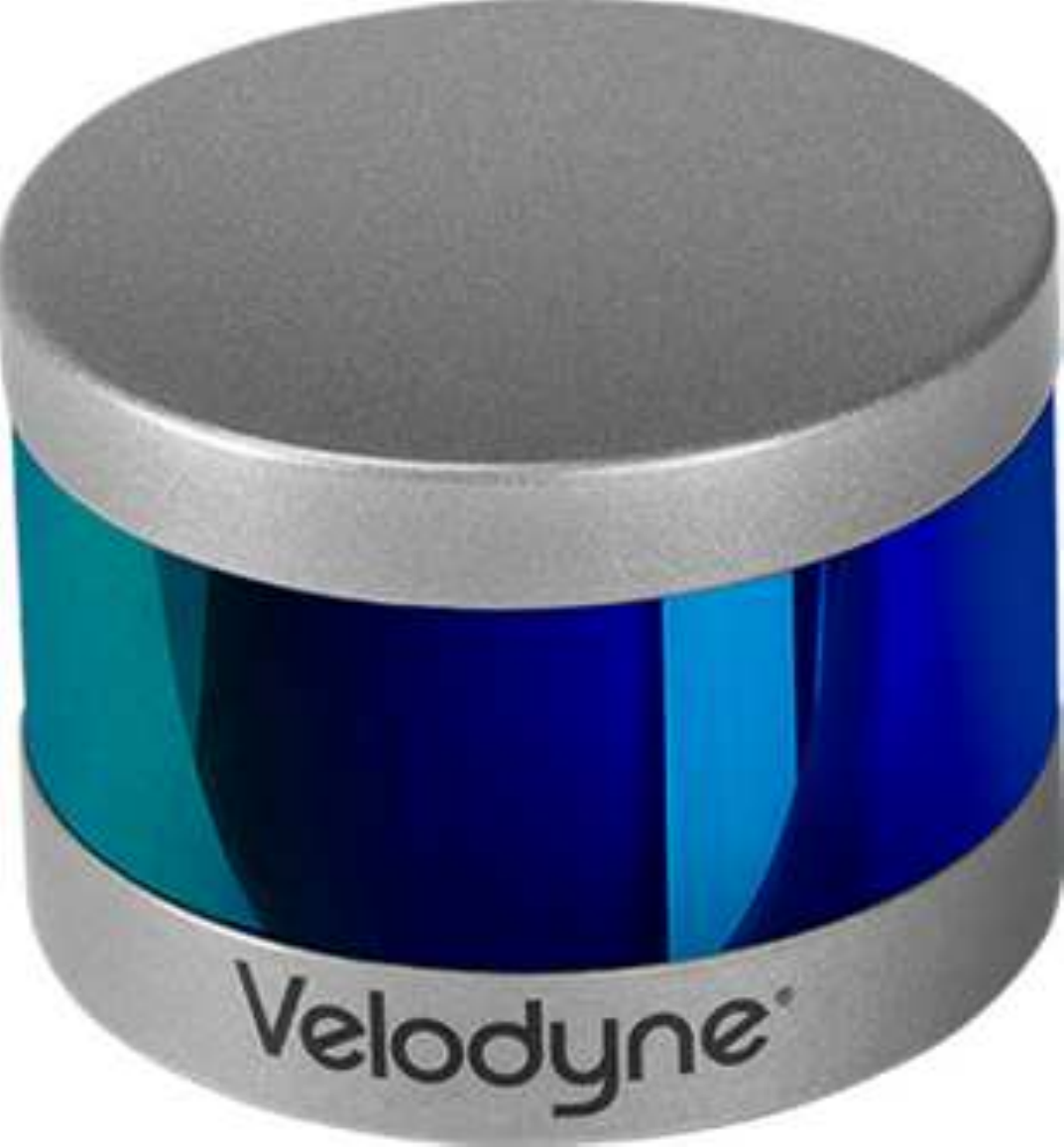}\\
					VLP-16\\\cite{vlp16}
				\end{minipage} &
				\begin{minipage}{0.04895\paperwidth}
					\raggedright
					\includegraphics[width=0.02926\paperwidth]{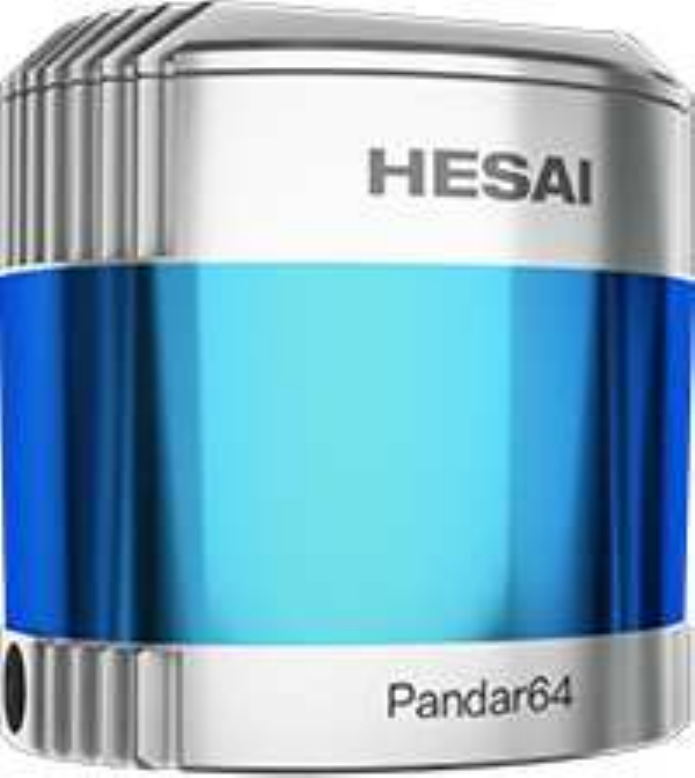}\\
					Pandar64\\\cite{pandar64}
				\end{minipage} &
				\begin{minipage}{0.06395\paperwidth}
					\raggedright
					\includegraphics[width=0.02926\paperwidth]{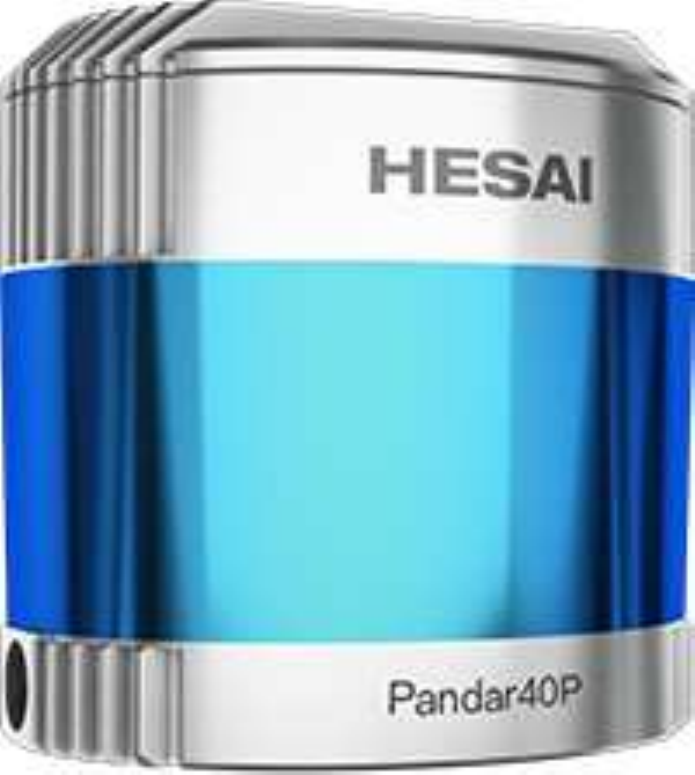}\\
					Pandar40P\\\cite{pandar40p}
				\end{minipage} &
				\begin{minipage}{0.06395\paperwidth}
					\raggedright
					\includegraphics[width=0.02144\paperwidth]{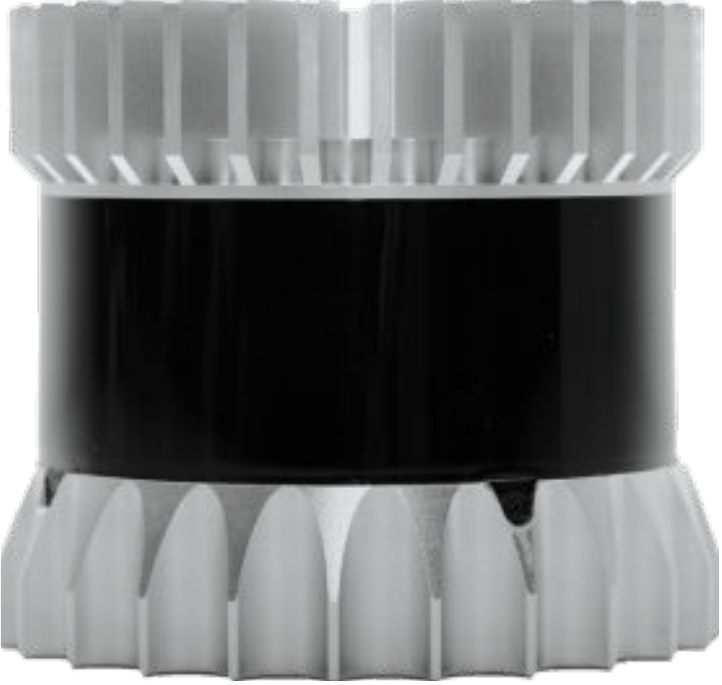}\\
					OS1-64\\\cite{os1}
				\end{minipage} &
				\begin{minipage}{0.06395\paperwidth}
					\raggedright
					\includegraphics[width=0.02144\paperwidth]{os-1-64.pdf}\\
					OS1-16\\\cite{os1}
				\end{minipage} &
				\begin{minipage}{0.06395\paperwidth}
					\raggedright
					\includegraphics[width=0.02875\paperwidth]{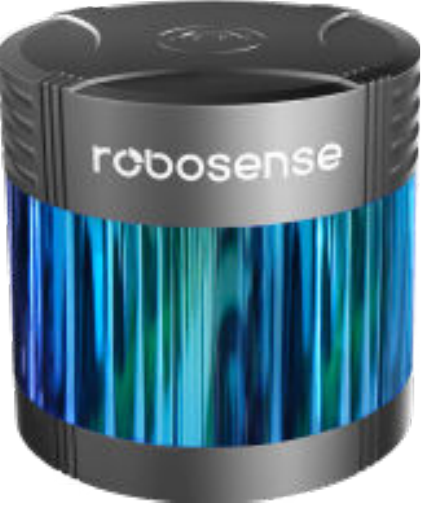}\\
					RS-Lidar32\\\cite{rslidar32}
				\end{minipage} \\ 
				\hline \T
				Channels & 128 & 64 & 32 & 32 & 16 & 64 & 40 & 64 & 16 & 32 \\ 	    
				FPS[Hz] & 5-20 & 5-20 & 5-20 & 5-20 & 5-20 & 10,20 & 10,20 & 10,20 & 10,20 & 5,10,20 \\
				Precision[m] & $\pm0.03$ & $\pm0.02^{\bm{a}}$ & $\pm0.02$ & $\pm0.03$ & $\pm 0.03$ & $\pm0.02^{\bm{c}}$ & $\pm 0.02^{\bm{c}}$ & $\pm0.03^{\bm{d}}$ & $\pm 0.03^{\bm{d}}$ & $\pm0.03^{\bm{c}}$ \\
				Max.Range[m] & $245$ & $120$ & $100$ & $200$ & $100$ & $200$ & $200$ & $120$ & $120$ & $200$ \\
				Min.Range[m] &  & $3$ & $2$ & $1$ & $1$ & $0.3$ & $0.3$ & $0.8$ & $0.8$ & $0.4$  \\ 
				VFOV[deg] & $40$ & $26.8$ & $41.33$ & $40$ & $30$ & $40$ & $40$ & $33.2$ & $33.2$ & $40$ \\
				VB[deg] & {\vfovbounds{15}{25}} & {\vfovbounds{2}{24.8}} & {\vfovbounds{10.67}{30.67}} & {\vfovbounds{15}{25}} & {\vfovbounds{15}{15}} & {\vfovbounds{15}{25}} & {\vfovbounds{15}{25}} & {\vfovbounds{16.6}{16.6}} & {\vfovbounds{16.6}{16.6}} & {\vfovbounds{15}{25}} \\
				HRes[deg] & $0.1$-$0.4$ & $0.09$ & $0.08$-$0.33$ & $0.1$-$0.4$ & $0.1$-$0.4$ & $0.2$,$0.4$ & $0.2$,$0.4$ & $0.7$,$0.35$,$0.17$ & $0.7$,$0.35$,$0.17$ & $0.1$-$0.4$ \\
				VRes[deg] & ${0.11}^{\bm{b}}$ & ${0.33}^{\bm{a}}$ & $1.33$ & ${0.33}^{\bm{b}}$ & $2.0$ & $0.167{}^{\bm{b}}$ & $0.33{}^{\bm{b}}$ & $0.53$ & $0.53$ & ${0.33}^{\bm{b}}$ \\
				$\lambda$[nm] & 903 & 903 & 903 & 903 & 903 & 905 & 905 & 850 & 850 & 905 \\
				$\phi$[mm] & 165.5 & 223.5 & 85.3 & 103 & 103.3 & 116 & 116 & 85 & 85 & 114 \\
				Weight(kg) & 3.5 & 13.5 & 1.0 & 0.925 & 0.830 & 1.52 & 1.52 & 0.425 & 0.425 & 1.17 \\
				Firmware ver. & ${}^{\bm{e}}$ & 4.07 & 2.1.7.1 & N/A & 3.0.29.0 & 5.10 & 4.29 & 1.12.0 & 1.12.0 & ${}^{\bm{f}}$ \\	
				\hline
			\end{tabular}
			\vspace{1pt}
			\caption{LiDARs tested in this study, by manufacturer and number of channels (rings).\protect\footnotemark Acronyms are frames per second (FPS), vertical field-of-view (VFOV), VFOV upper and lower bounds (VB), horizontal resolution (HRes), vertical resolution (VRes), laser wavelength ($\lambda$), and diameter $\phi$.
				$\,^{\bm{*}}$Velodyne VLS128 pre-production model (63-9480 Rev-3). 
				$\,^{\bm{a}}$Velodyne states HDL-64S2 accuracy is $\pm2cm$ for 80\% of channels, and $\pm5cm$ for the remaining; VRes for $+2\degree\isep-8.33\degree$ is $1/3\degree$ and for $-8.83\degree\isep-24.33\degree$ is $1/2\degree$. 
				$\,^{\bm{b}}$Minimum (or finest) resolution, as these sensors have variable angle difference between beams. 
				$\,^{\bm{c}}$Hesai and RoboSense state that accuracy for $0.3m\isep 0.5m$ is $\pm0.05$m, then $\pm0.02$m from $0.5m\isep 200m$. 
				$\,^{\bm{d}}$Ouster states accuracy for $0.8m\isep 2m$ is $\pm0.03m$, for $2m\isep 20m$ is $\pm0.015m$, for $20m\isep 60m$ is $\pm0.03m$, and over $60m$ is $\pm0.10m$.
				${}^{\mathbf{e}}$VLS-128 firmware is not stated here as it was not a production model.
				${}^{\mathbf{f}}$RS-Lidar32 had top board firmware version T9R23Va\_Tb\_00 and bottom board firmware version B8R02Va\_T5\_A.}
			\label{tab:lidar-list}
			\vspace{-2.5em}
		\end{center}
	\end{table*}
}

\section{LiDAR datasets}
\label{s:relworks}
Table~\ref{tab:datasets} summarizes current datasets featuring LiDARs, and highlights the contributions made by our dataset. The Stanford Track Collection\cite{stanford2011} carefully records tracks of objects and their dataset offer the object tracks, while FORD Campus vision and LiDAR dataset\cite{ford2011} include several complete scenes captured by multiple LiDARs. The Oxford RobotCar Dataset\cite{RobotCarDatasetIJRR} has one 3D LiDAR and two 2D LiDARs, and accumulation of 2D data as the vehicle moves allows the reconstruction of 3D scenes. ApolloScape\cite{apolloscape2018} features two 3D LiDARs, in several environments, times of the day and varying weather. The KAIST dataset\cite{kaist2018} features 3D LiDAR (905\,nm infrared) plus normal vision and a thermal (long wave infrared 8\,$\mu$m to 15\,$\mu$m), and is therefore considered multispectral. The Lidar-video driving dataset\cite{li-vi2018} also collects data from one LiDAR, a camera and CAN bus data targeting driving behaviour.

More recently, the ArgoVerse dataset\cite{argoverse} features two LiDARs, one on top of the other, plus a ring of cameras for 360$\degree$ annotation. Vector maps (HD maps) are also provided. The nuScenes dataset by Aptiv\cite{nuscenes2019} features one LiDAR, several cameras and other sensors, and is captured in a diverse range of environments, times of day and weather conditions. The Honda Research Institute 3D dataset (H3D)\cite{h3d} also features one LiDAR and multiple sensors, with labels provided at 2\,Hz and propagated at 10\,Hz so as to provide labels at the same rate as the LiDAR. Similarly, Lyft dataset\cite{lyft2019} features 3 LiDARs and an array of cameras, and different versions of the dataset are available. The Waymo Open Dataset\cite{waymo2019} features 5 LiDARs created by Google/Waymo, one 360$\degree$ and 4 for lower FOV and proximity detection 
in several different locations. The A2D2 dataset by Audi\cite{geyer2020a2d2} features 5 VLP-16 LiDARs tilted to cover the immediate surroundings of the vehicle in diverse environments. 
\footnotetext{Sensor images are not to scale and copyrights are owned by their respective manufacturers.}

Different from the above works, this would be the first dataset to collect data under the similar conditions but with different LiDARs. Some of the above datasets feature more than one LiDAR but with limited models, while in our work we offer {\numlidars} different models. Also, as far as we know, no static tests of LiDARs are publicly available. 

Besides datasets featuring LiDARs, other related works have consider diverse LiDAR evaluations. Jokela\etal\cite{jokela2019testing} tested 5 different LiDARs  
in fog and rain conditions at Clermont-Ferrand's 31\,m long fog chamber\cite{colomb2008innovative}, including different perception targets and conditions; they also evaluated these LiDARs under low temperature snowy environments. The EU project DENSE\cite{dense2019,ritter2019dense} tested 2 different LiDARs plus gated camera, FIR camera and other devices under adverse weather conditions in urban environments and also used the Clermont-Ferrand fog chamber.
While our present study currently lacks evaluations under snowy conditions, we test a broader range of sensors in a wider variety of adverse weather experiments.

{
	\begin{figure*}[!htb]
		\centering
		\subfloat[][]{
			\includegraphics[width=0.38\textwidth]{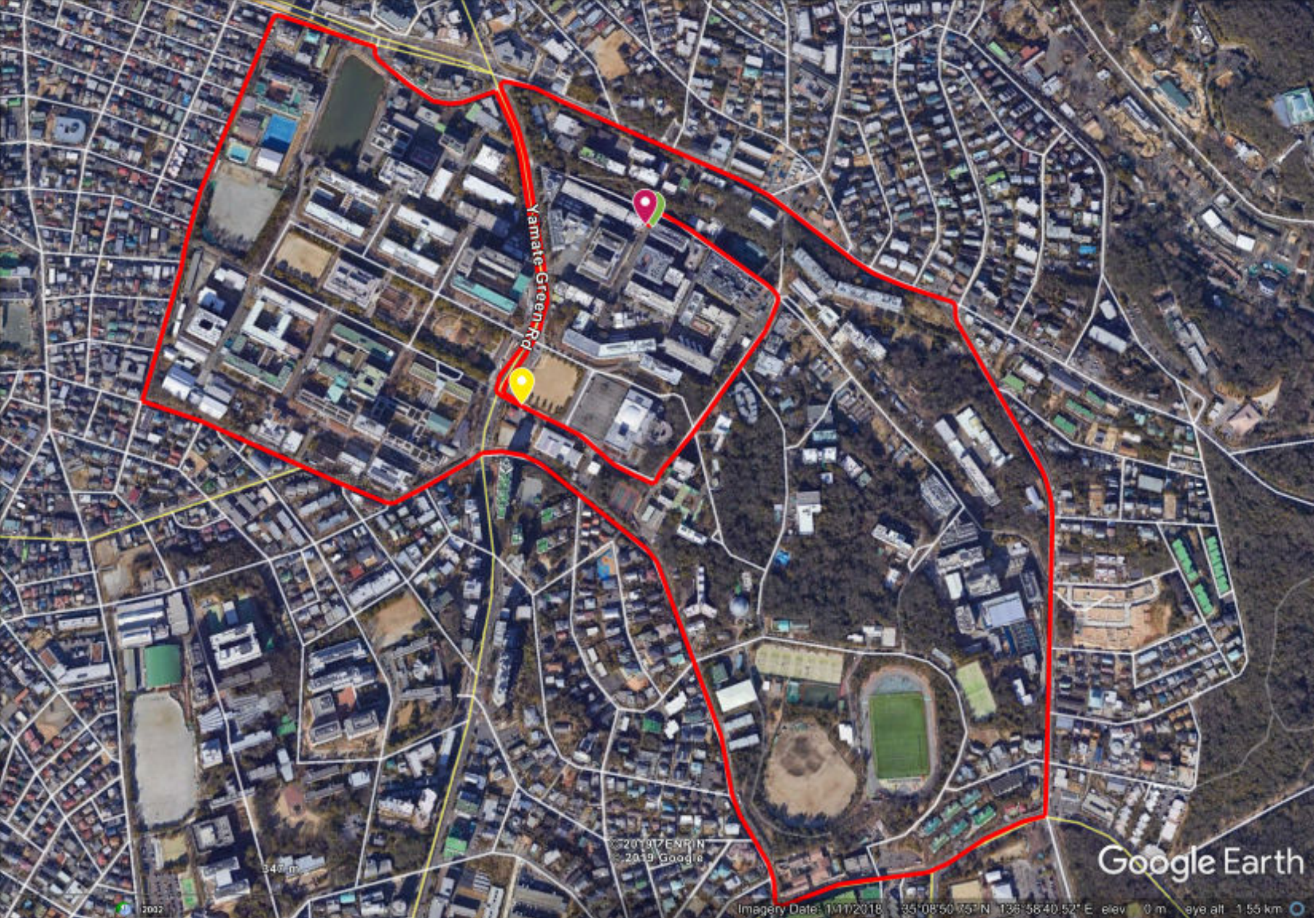}
			\label{F:m3ldmap-a}
		}
		\subfloat[][]{
			\includegraphics[width=0.265\textwidth]{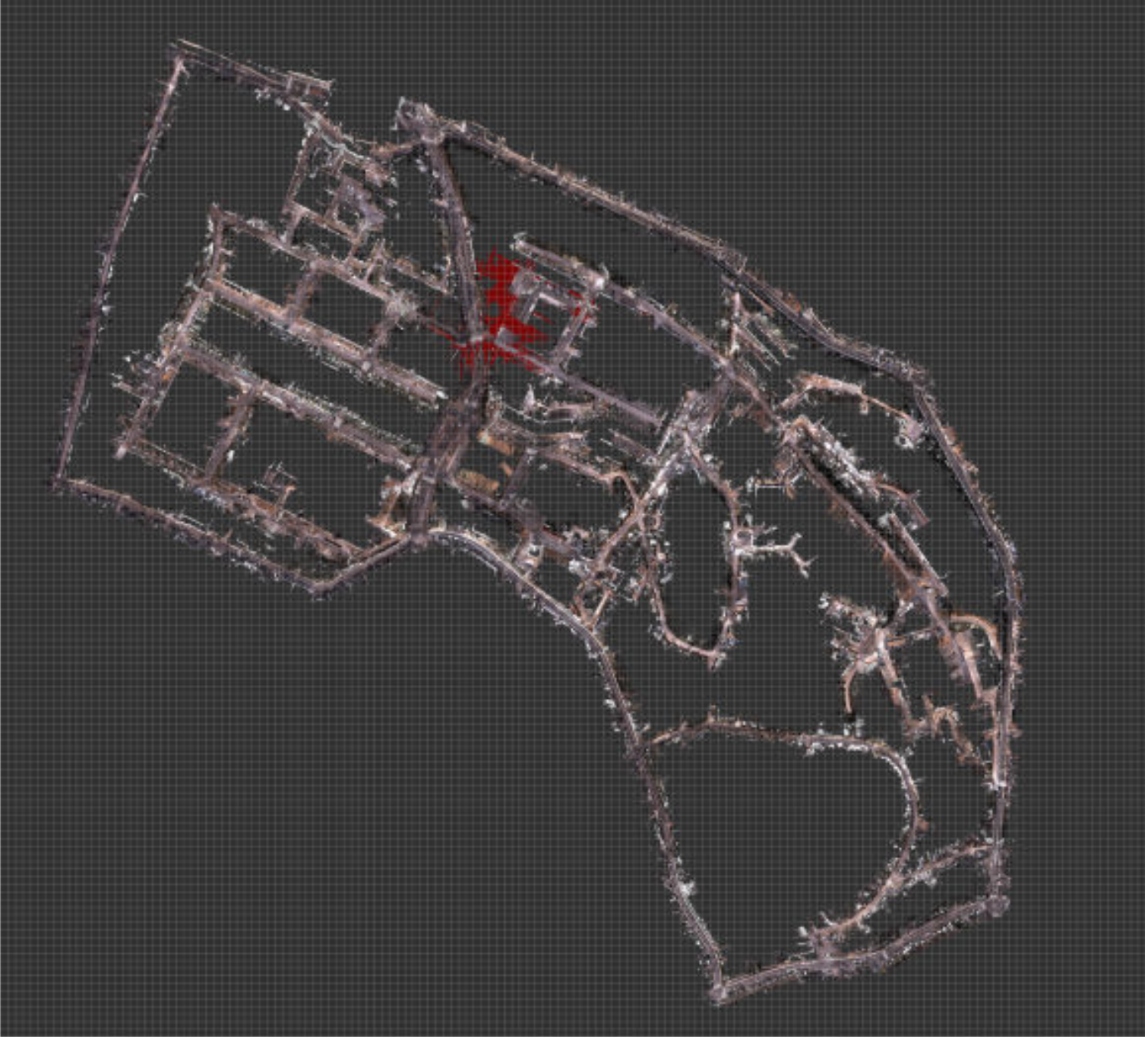}
			\label{F:m3ldmap-b}
		}
		\subfloat[][]{
			\begin{minipage}[b]{.295\textwidth}
				\includegraphics[width=\textwidth]{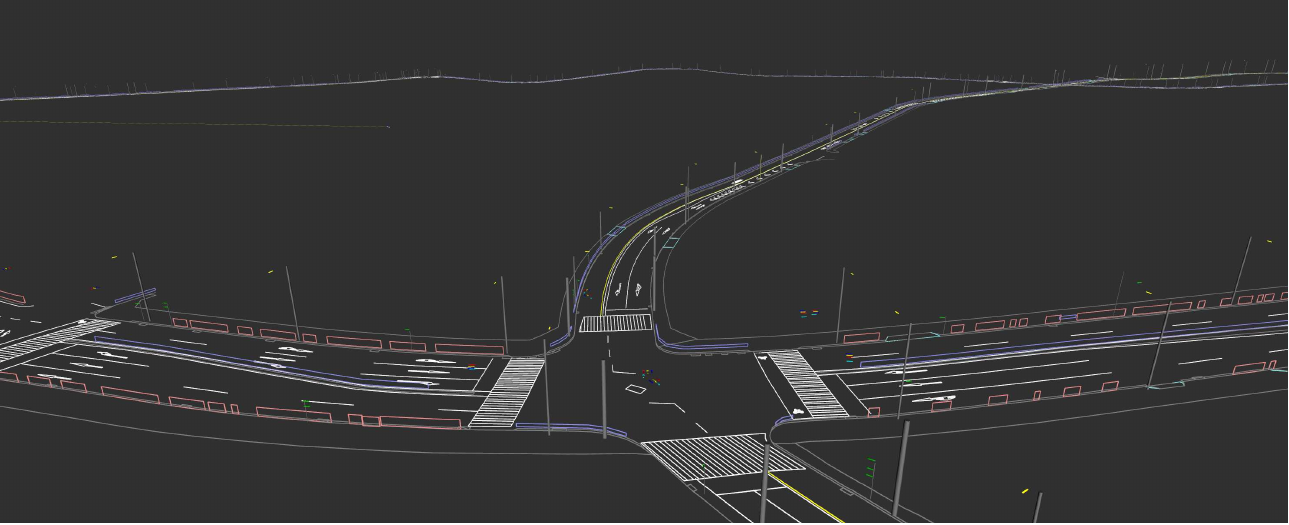}
				\vfill
				\includegraphics[width=\textwidth]{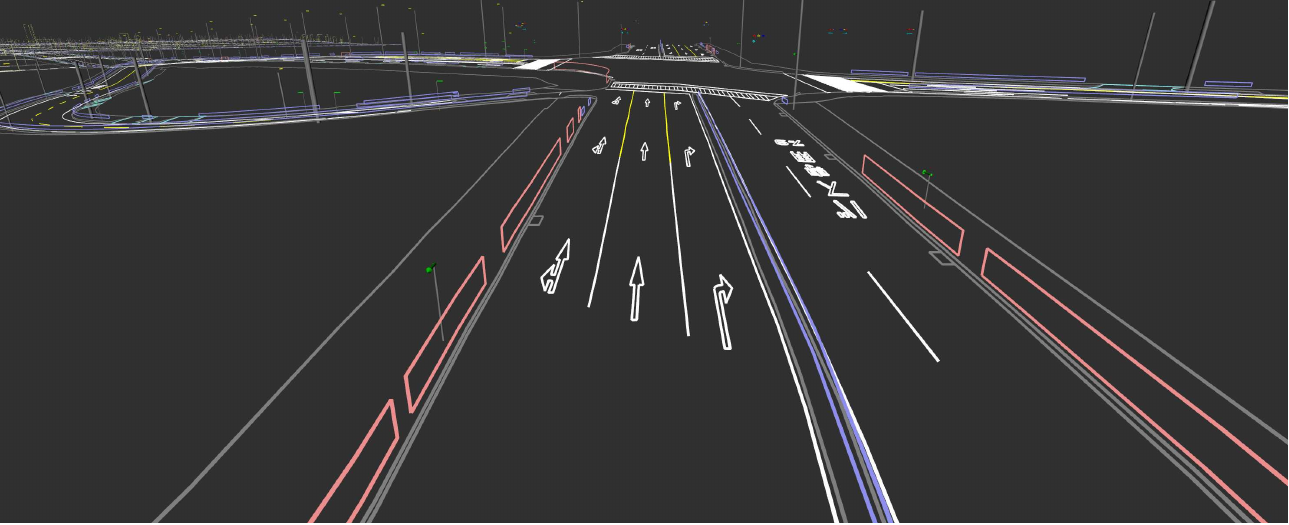}
			\end{minipage}
			\label{F:m3ldmap-c}
		}
		\caption[]%
		{Map of the dynamic environment included in the dataset: \subref{F:m3ldmap-a} location reference, followed trajectory is shown in red (total length of 6.56\,km),
			{\color{red}{\faMapMarker}} and {\color{green}{\faMapMarker}} markers denote the starting and goal points, respectively, {\color{yellow}{\faMapMarker}} corresponds to a vehicle gate in/out the campus.
			\subref{F:m3ldmap-b} is the pointcloud map (grid cell size 10\,m) and \subref{F:m3ldmap-c} some scenes with the vector map.}
		\label{F:m3ldmap}
		\vspace{-1em}
	\end{figure*}
}

\begin{figure}[!htb]
	\centering
	\includegraphics[width=0.46\textwidth]{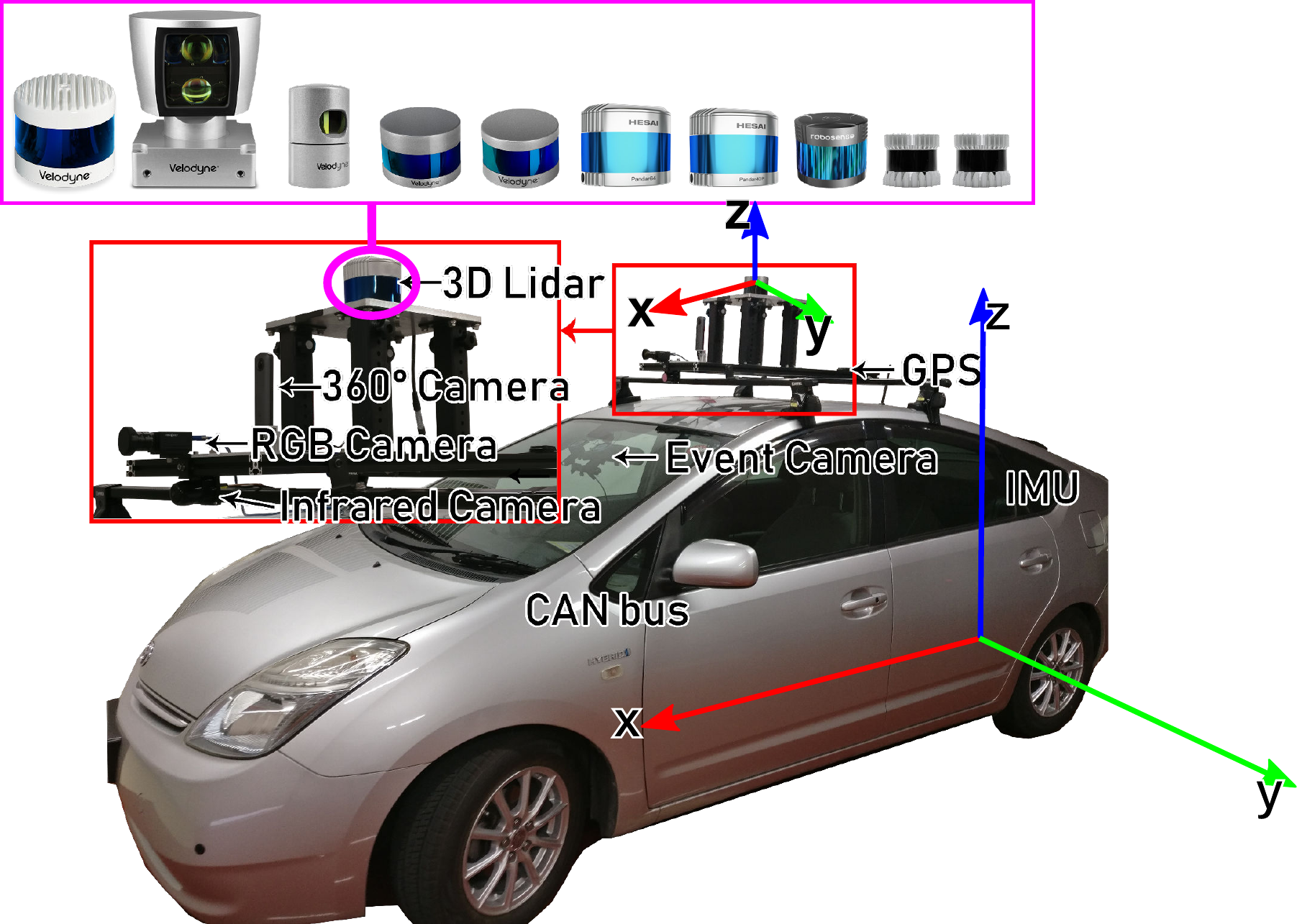}
	\caption{Instrumented vehicle used to capture static and dynamic data, sensors are mounted on a metal plate about 2\,m from the ground.}
	\label{F:ginpuri}
	\vspace{-1em}
\end{figure}

\section{{\datasetname} Dataset}
\label{s:dataset}
The {\datasetname} dataset features 5 LiDARs from Velodyne Lidar\footnote{\url{https://velodynelidar.com}}, two from Ouster Inc.\footnote{\url{https://ouster.com}}, two from Hesai Photonics Technology Co., Ltd\footnote{\url{https://www.hesaitech.com}}, and one from RoboSense--Suteng Innovation Technology Co., Ltd.\footnote{\url{http://www.robosense.ai}}. 
Table~\ref{tab:lidar-list} describes the general specifications of each tested device. Data from all LiDAR and all environments was collected from April to September 2019. 

All sensors tested in this study were off-the-shelf production models with the exception of the Velodyne VLS-128. This sensor was a pre-production model, and was tested to provide a preview for the production 128-line Alpha Prime sensor which was unavailable at the time the experiments were carried out. The dataset will be extended with the Alpha Prime results when testing has been completed.

All these sensors correspond to the multi-beam (multi-channel) mechanical scanning type: several pairs of laser diodes and photo-detectors (avalanche photo detector (APD) and single-photon avalanche diode (SPAD)) and corresponding emit-remit optics and mirrors, are rotated by a motor for 360$\degree$ which defines azimuth, while the vertical angle of a laser and photo-detector pair defines elevation. All sensors in this selection have short-wave infrared (SWIR) wavelengths between 850\,nm, 903\,nm and 905\,nm. While some support multiple returns (echoes), the data collected in our dataset always records only the strongest echo.

\section{Dynamic data}
\label{s:env-dynamic}

\subsection{Data Collection}
\label{ss:datacollection}
The target was to collect data in a variety of traffic conditions, including different types of environments, varying density of traffic and times of the day. We drove our instrumented vehicle, a Toyota Prius shown in Fig.~\ref{F:ginpuri}, three times per day, around the trajectory shown in Fig.~\ref{F:m3ldmap}, and collected data for the following key time periods:
\begin{itemize}
	\item Morning (9am-10am)
	\begin{itemize}
		\item Pedestrian traffic: medium-low
		\item Vehicle traffic: high
		\item Conditions: people commuting, students and staff arriving on the campus. Clear to overcast weather.
	\end{itemize}
	\item Noon (12pm-1pm)
	\begin{itemize}
		\item Pedestrian traffic: high
		\item Vehicle traffic: medium-low
		\item Conditions: large number of students and staff heading to and from cafeterias and restaurants. Clear to overcast weather.
	\end{itemize}
	\item Afternoon (2pm-4pm)
	\begin{itemize}
		\item Pedestrian traffic: low
		\item Vehicle traffic: medium-low
		\item Conditions: busy work and class period. Clear to overcast weather.
	\end{itemize}
\end{itemize}
Fig.~\ref{F:ginpuri} shows the vehicle used for data capture. The 3D LiDAR on top was replaced for each experiment only after the three time periods were recorded, and only one LiDAR was used at a time to avoid noise due to mutual interference. Data from other sensors (RGB camera, IR camera, 360$\degree$ camera, event camera, IMU, GNSS, CAN) was recorded together with the LiDAR data, every data with corresponding timestamps, using ROS\cite{ros2009}. In addition, we collected calibration data for each new LiDAR setup to perform extrinsic LiDAR to camera calibration, using a checkerboard and various other points of interest. Clear lighting conditions were ensured when recording such data. 

The routes driven in this data capture also have a reference pointcloud map available, which was created by a professional mobile mapping system (MMS). This map includes RGB data, and vector map files (HD map) for public road outside of the Nagoya University campus, and is also provided as part of the dynamic traffic data.
\begin{figure}[!htb]
	\centering
	\includegraphics[width=0.45\textwidth]{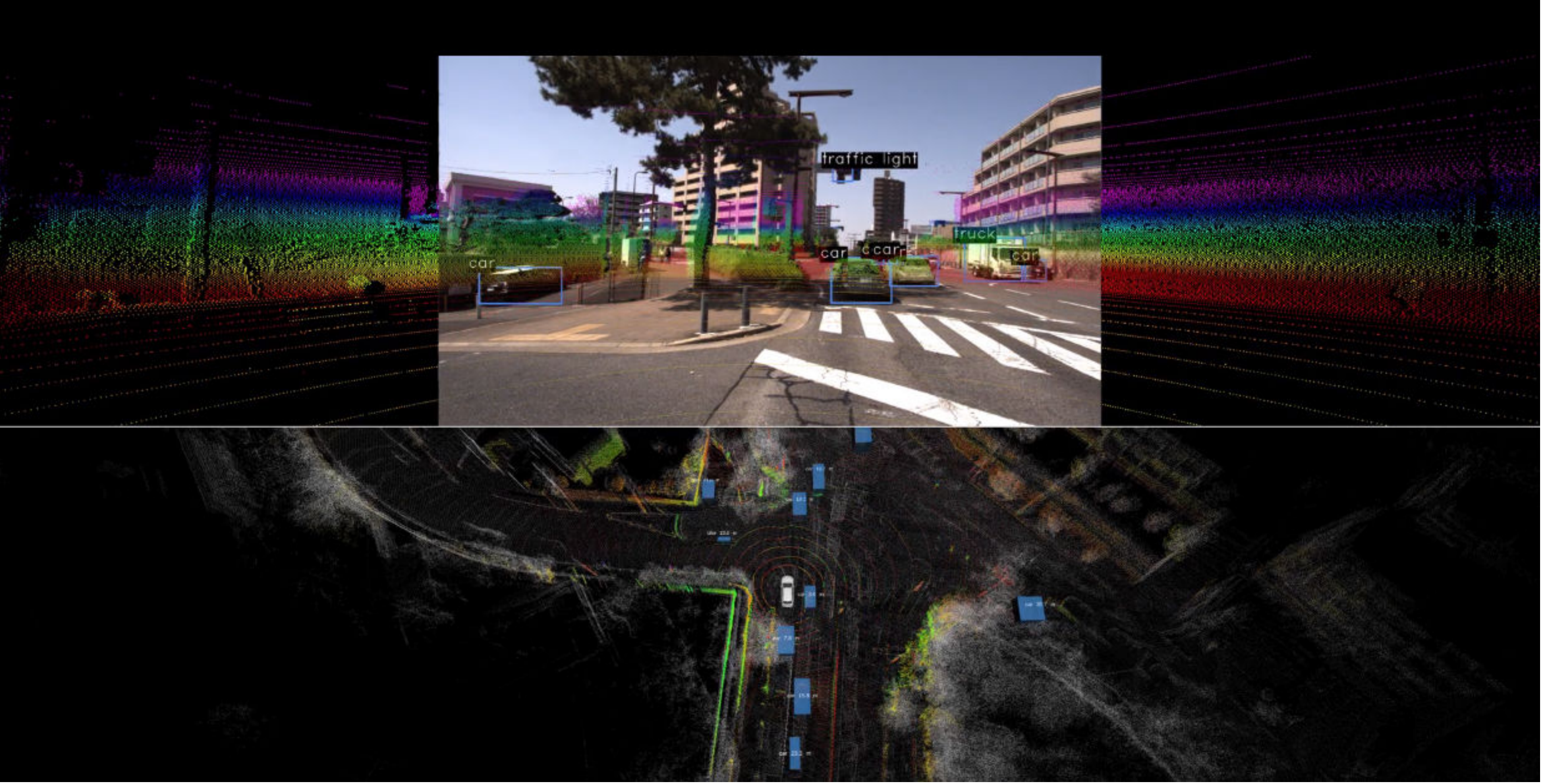}
	\caption{Dynamic traffic scenes by applying SOTA algorithms on pointcloud.}
	\label{F:dynamic}
	\vspace{-1em}
\end{figure}

\begin{figure*}[!htb]
	\centering
	\sbox{\arrangebox}{%
		\subfloat[][]{
			\centering
			\includegraphics[width=.3\textwidth,height=.36\textwidth]{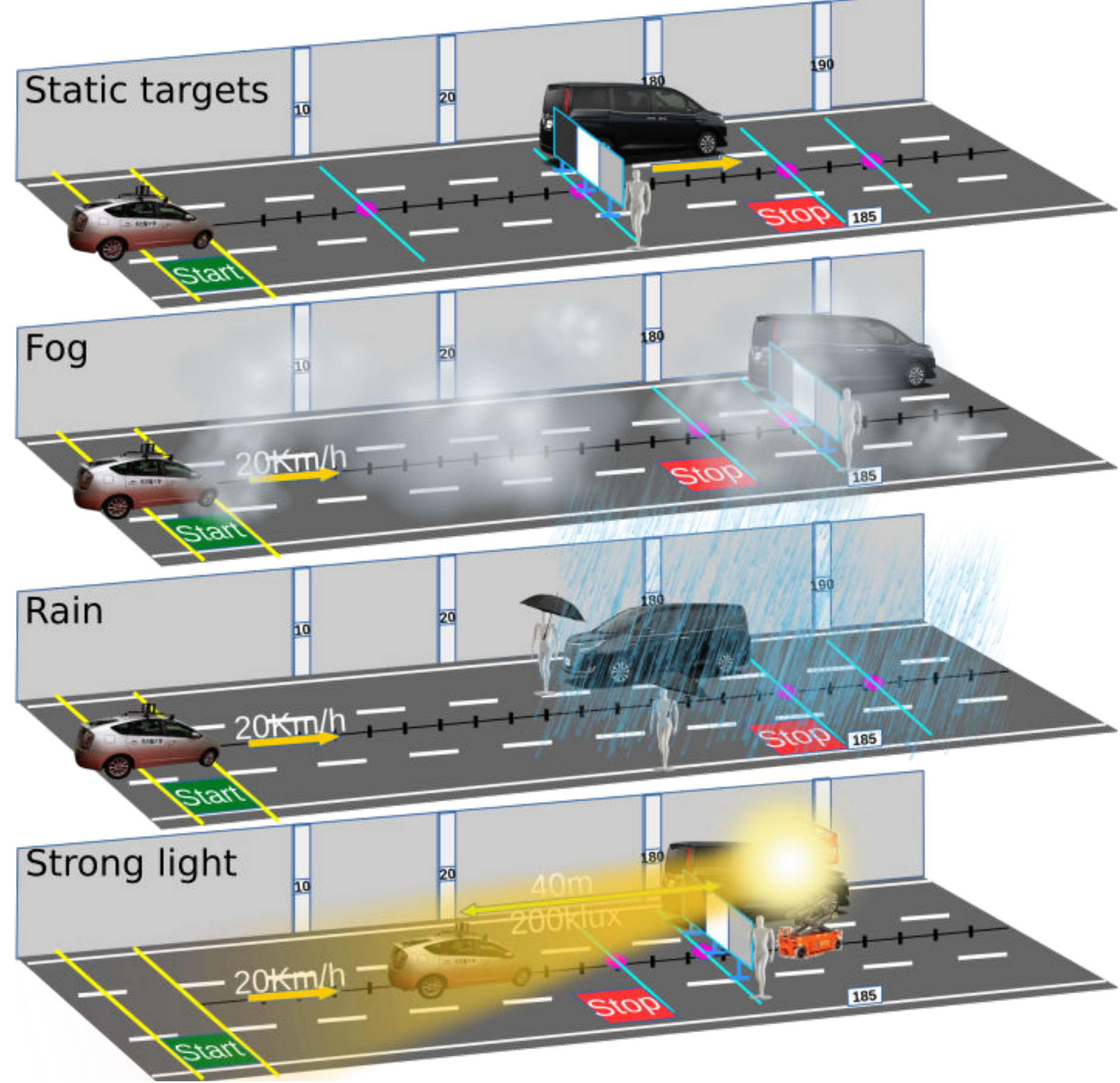}
			\label{F:jaritargets-a}
		}%
	}
	\setlength{\arrangeht}{\ht\arrangebox}
	\usebox{\arrangebox}\hspace{-1em}
	\begin{minipage}[b][\arrangeht][s]{0.7\textwidth}
		\begin{center}
			\centering
			\subfloat[][]{
				\includegraphics[width=0.27\textwidth]{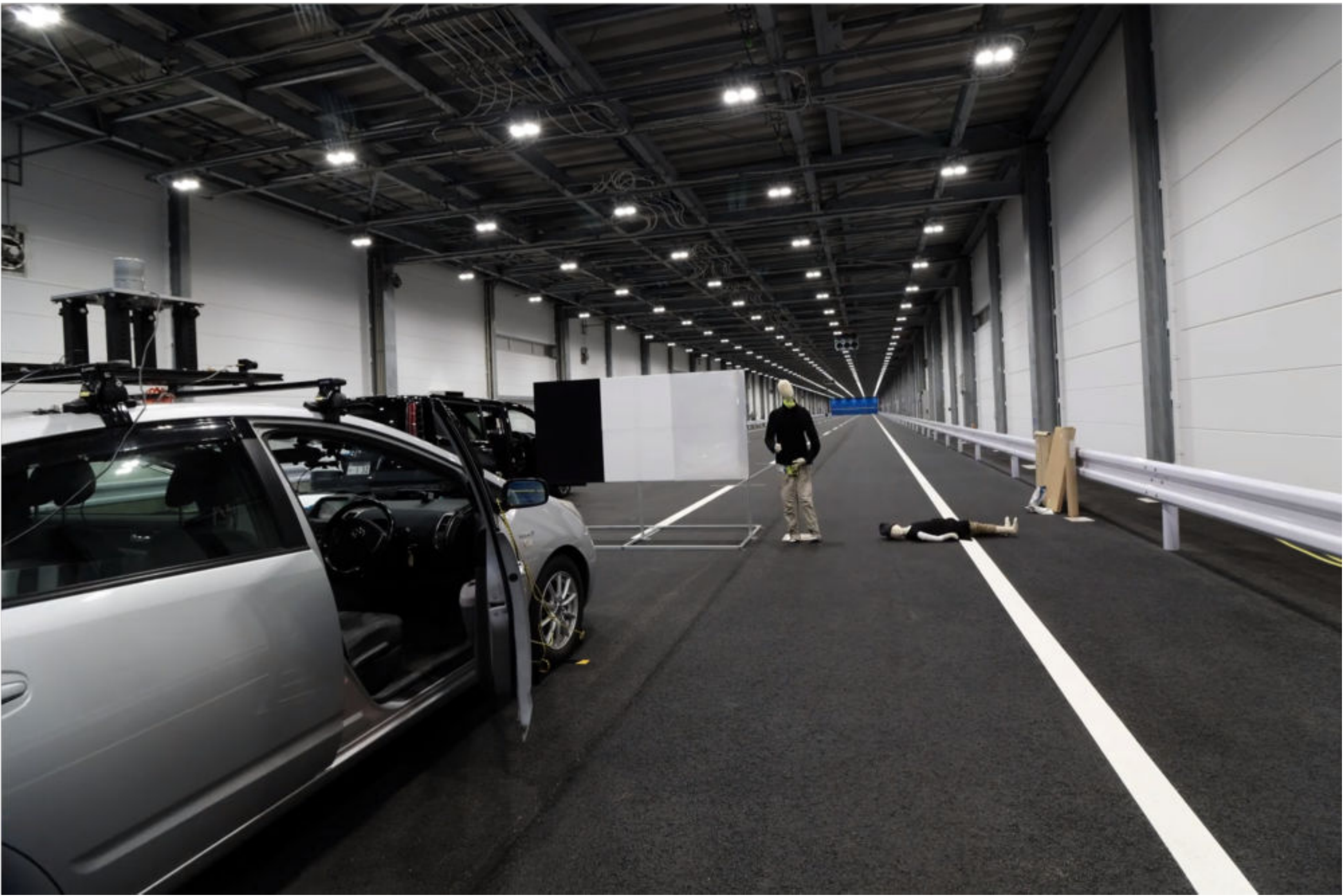}
				\label{F:jaritargets-b}
			}\hspace{-10pt}
			\subfloat[][]{
				\includegraphics[width=0.4\textwidth]{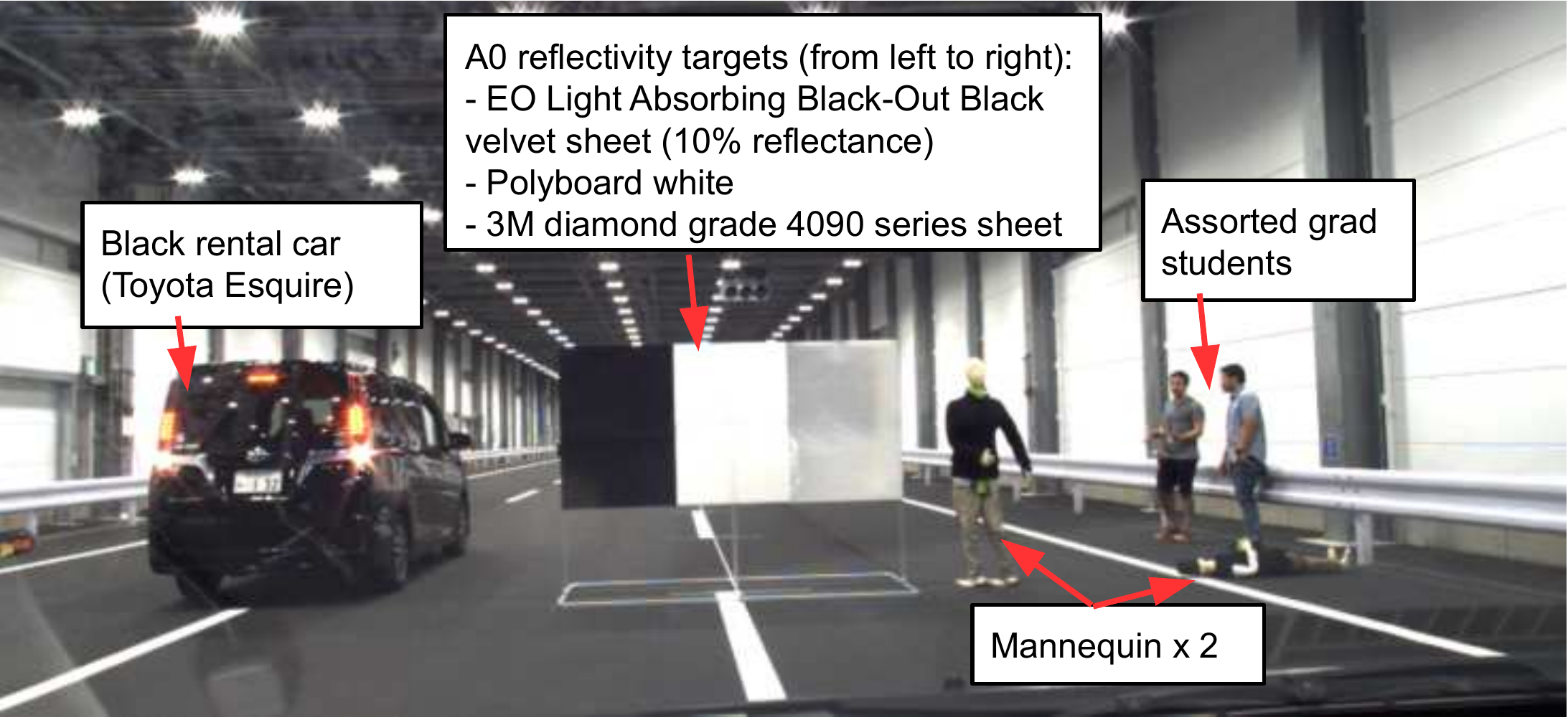}
				\label{F:jaritargets-c}
			}\hspace{-10pt}
			\subfloat[][]{
				\includegraphics[width=0.27\textwidth]{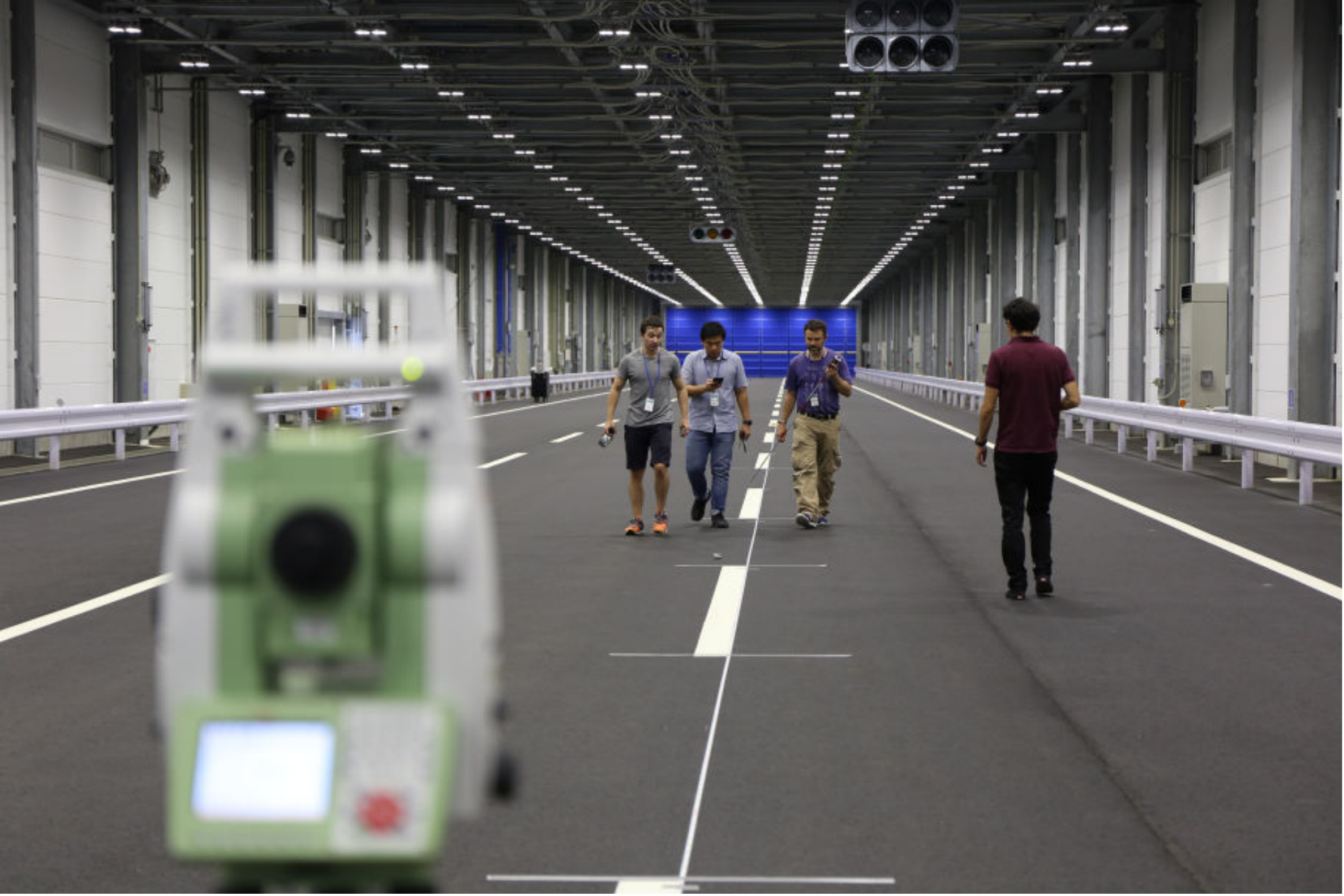}
				\label{F:jaritargets-g}
			}\\
			\subfloat[][]{
				\includegraphics[width=0.31\textwidth]{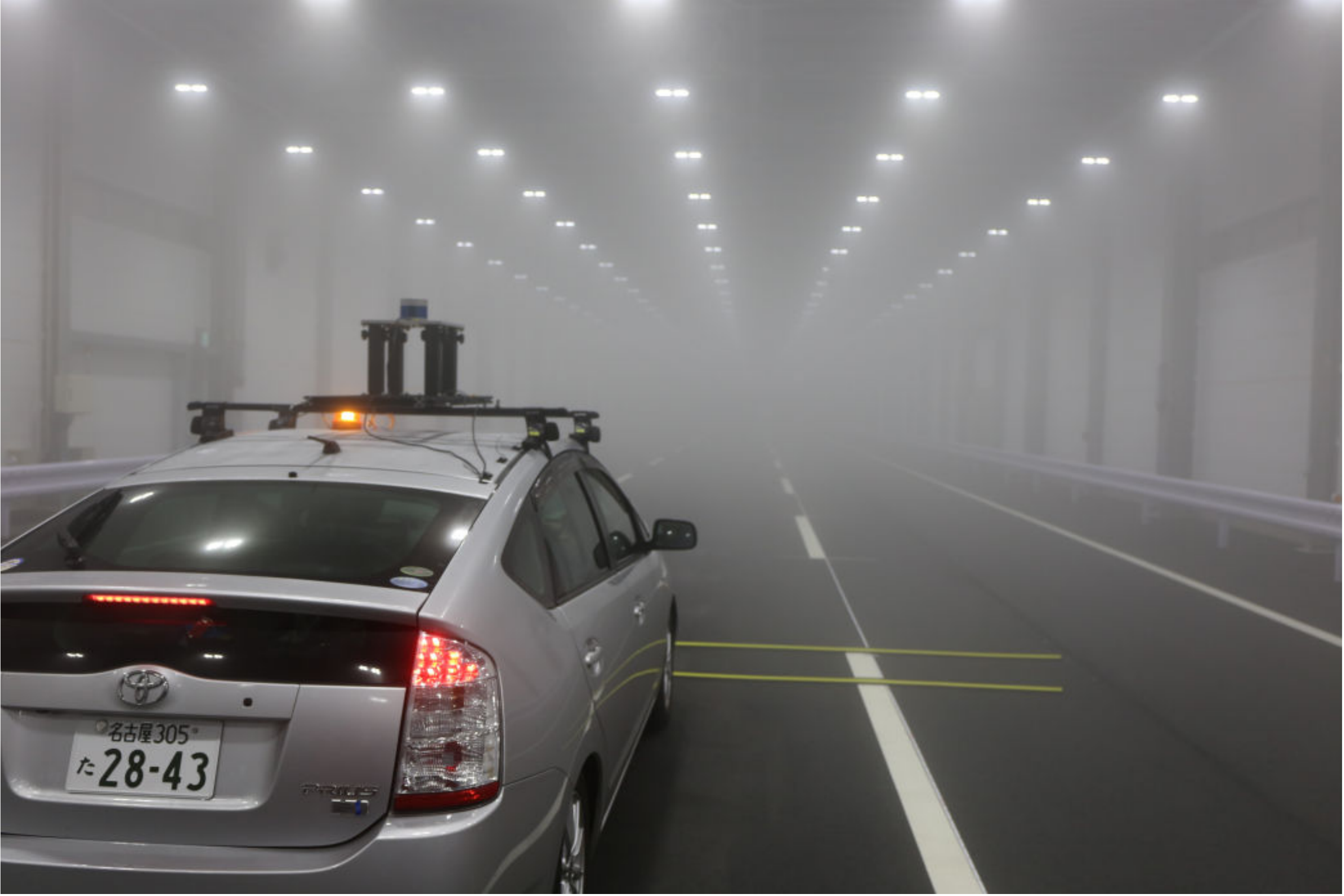}
				\label{F:jaritargets-d}
			}\hspace{-10pt}
			\subfloat[][]{
				\includegraphics[width=0.31\textwidth]{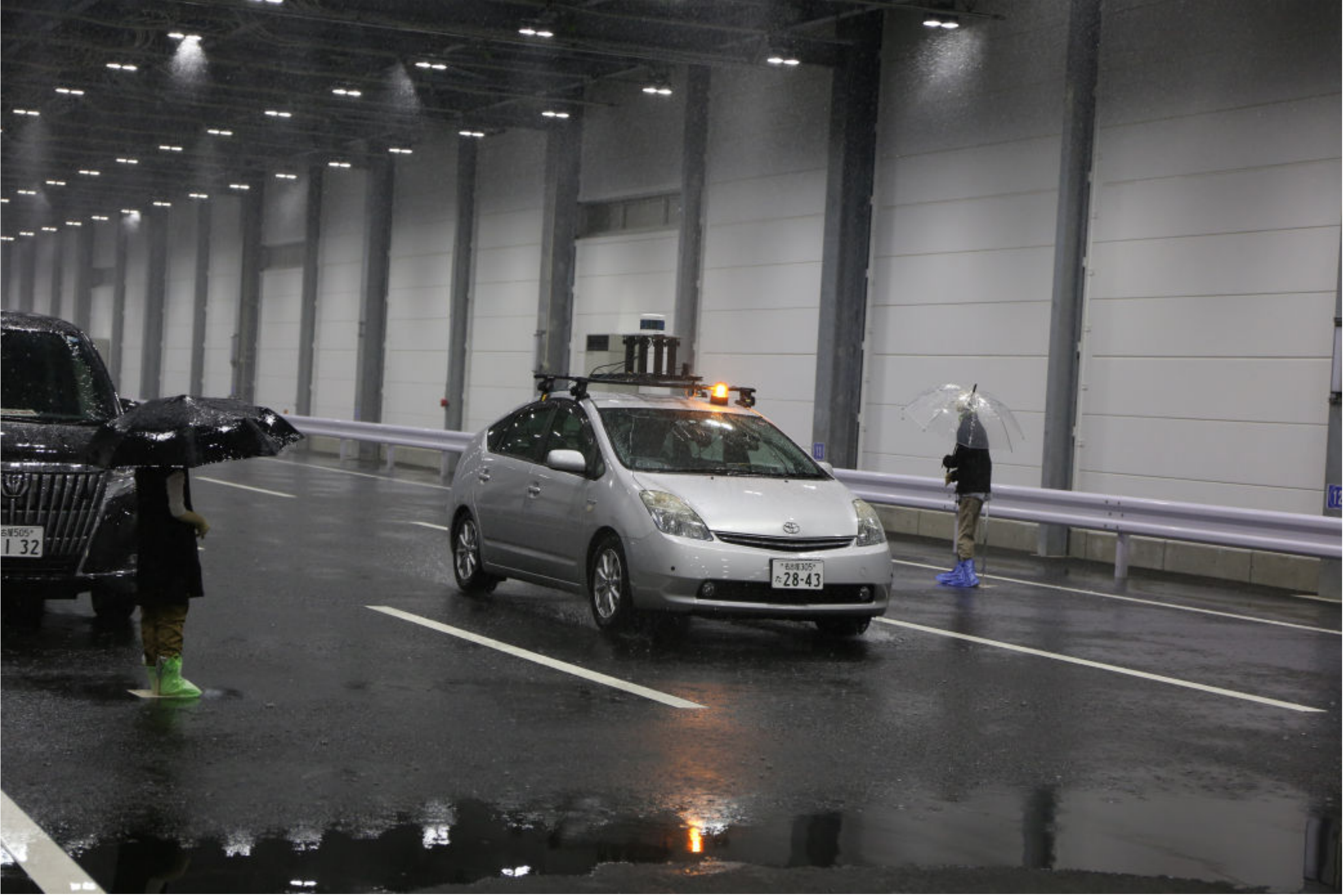}
				\label{F:jaritargets-e}
			}\hspace{-10pt}
			\subfloat[][]{
				\includegraphics[width=0.31\textwidth]{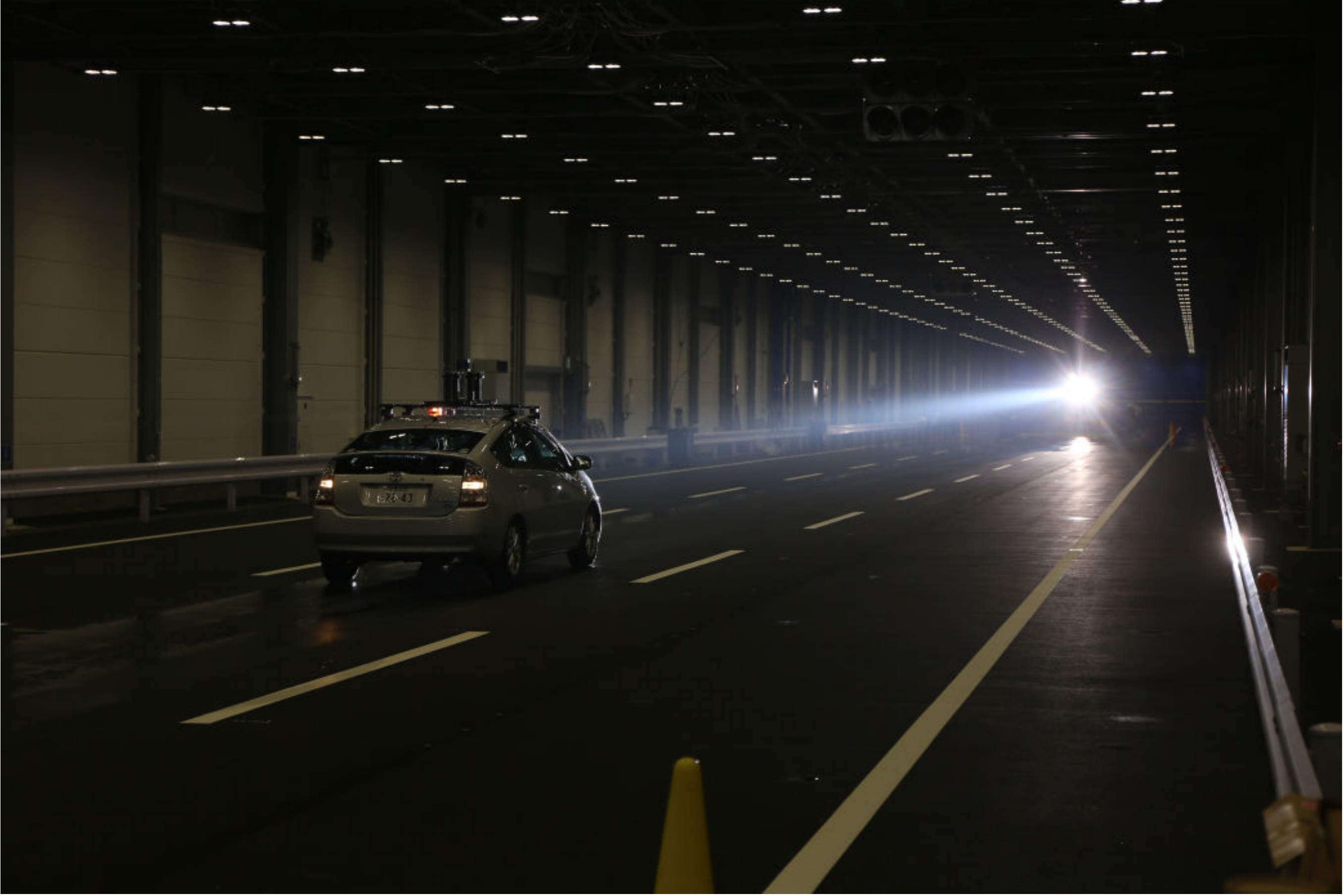}
				\label{F:jaritargets-f}
			}
		\end{center}
	\end{minipage}	
	\caption{Static targets and adverse weather experiments at JARI's weather chamber: \subref{F:jaritargets-a} configuration of the different scenarios,  \subref{F:jaritargets-b} and \subref{F:jaritargets-c} measurement, \subref{F:jaritargets-d} to \subref{F:jaritargets-f} sample adverse weather scenes, \subref{F:jaritargets-g} setting up ground truth.}
	\label{F:jaritargets}
	\vspace{-1em}
\end{figure*}
\begin{figure*}[!htb]
	\centering
	\includegraphics[width=0.8\textwidth]{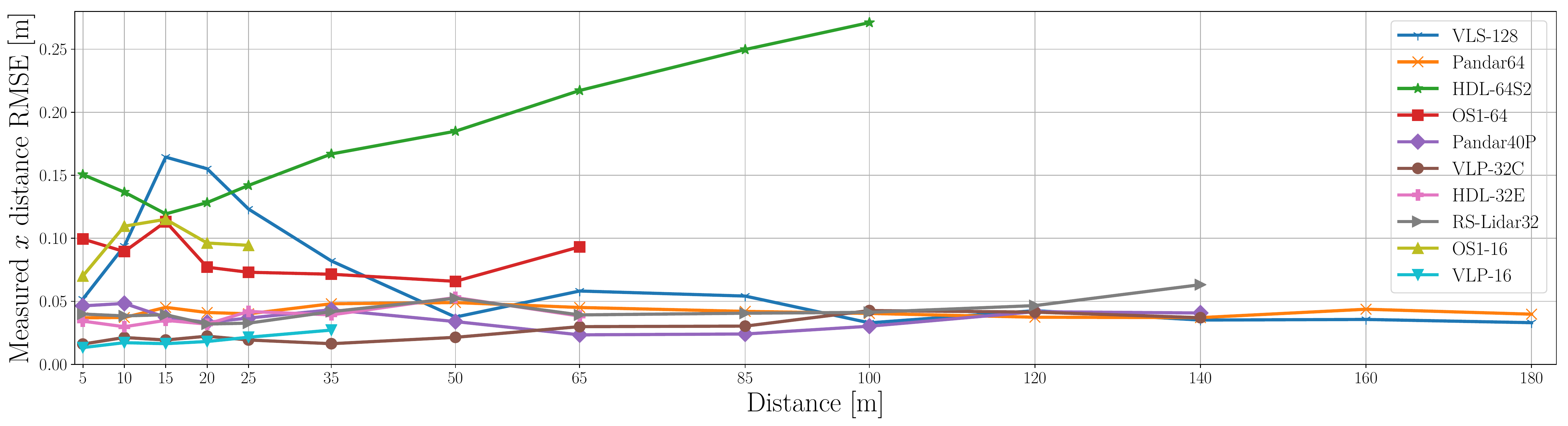}
	\caption{Range RMSE on $x$-axis distance per LiDAR.}
	\label{F:range}
	\vspace{-1.5em}
\end{figure*}

\subsection{Evaluation in Autoware}
\label{ss:evalinautoware}
Fig.~\ref{F:dynamic} shows qualitative results of running state-of-the-art algorithms, implemented in the self-driving open source platform Autoware\footnote{https://gitlab.com/autowarefoundation/autoware.ai} (see Kato\etal\cite{kato2018}), on LiDAR pointclouds. Fig.~\ref{F:dynamic} shows VLS-128 pointcloud when localized using the Normal Distributions Transform (NDT), LiDAR/camera fusion, and CNN-based object detection.

\section{Static targets}
\label{s:env-static}
For the static targets and the adverse weather conditions, we used the Japan Automobile Research Institute (JARI\footnote{\url{http://www.jari.or.jp}}) weather experimental facilities. Fig.~\ref{F:jaritargets}\subref{F:jaritargets-a} shows a cross view of such facilities during our experiments. It is a 200\,m long and 15\,m wide indoor weather chamber with 3 straight and marked lanes (each 3.5\,ｍ wide as per Japanese regulations), regularly flat, with fences, traffic lights, controlled illumination and ventilation, and multiple sprinklers for fog and rain. A description of JARI's weather chamber equipment and conditions is given in Section~\ref{s:env-weather}. 

As shown on Fig.~\ref{F:jaritargets}\subref{F:jaritargets-c}, the static targets in this study include: A0 size (841\,mm x 1189\,mm) reflective targets (Edmund Optics light absorbing black-out black velvet sheet (10\% reflectance), polyboard white, and 3M diamond-grade 4090 series sheet), a Toyota Esquire black mini-van, two mannequins wearing black clothes, and occasionally human participants when conditions were safe. Reflective targets were fixed on an aluminum frame reinforced to prevent warping and with backing material to ensure sheets remained flat.  

\begin{table}
	\centering
	\begin{tabular}{c l | c l}
		\hline\hline 
		Distance     & Target Ground & Distance & Target Ground\\
		to Lidar [m] & Truth [m]     & to Lidar [m] & Truth [m]\\
		\hline
		5	&	4.982     & 65	&	65.008\\
		10	&	9.998     & 85	&	85.005\\
		15	&	14.994    & 100	&	100.010\\
		20	&	20.001    & 120	&	120.006\\
		25	&	25.999    & 140	&	140.005\\
		35	&	35.007    & 160	&	160.007\\ 
		50	&	49.997    & 180	&	180.007\\
		\hline
	\end{tabular}
	\caption{Target distances and LiDAR to targets ground truth, as measured by the TS15.}
	\label{tab:truth}
	\vspace{-2.5em}
\end{table}
During this experiment, each LiDAR was warmed up for at least 30\,min to increase detection accuracy of the photo-detectors. As shown in Fig.~\ref{F:jaritargets}\subref{F:jaritargets-g}, we used a Leica Geosystems Total Station Viva TS15\cite{leica-ts15} and reflector prisms to setup the ground truth for target positions. Table~\ref{tab:truth} shows the target distances (along the LiDAR's $x$-axis) and the actual measured distances with the TS15. Reflective targets were carefully aligned at each measurement position, which we previously marked on the road surface, while the mini-van and the mannequins were approximately aligned with this. Fig.~\ref{F:jaritargets}\subref{F:jaritargets-b} shows the 5\,m mark as an example.
\begin{figure*}[!htb]
	\centering
	\subfloat[expected][Expected]{
		\includegraphics[width=0.8\textwidth]{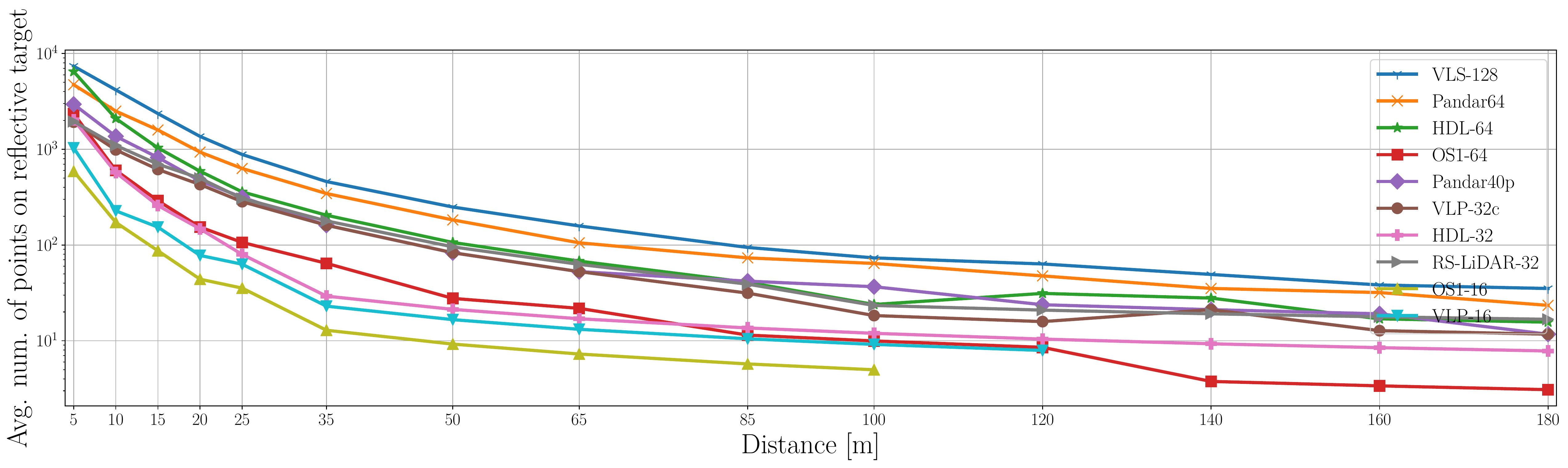}
		\label{F:density-a}
	}\\
	\vspace{-1em}
	\subfloat[actual][Measured]{
		\includegraphics[width=0.8\textwidth]{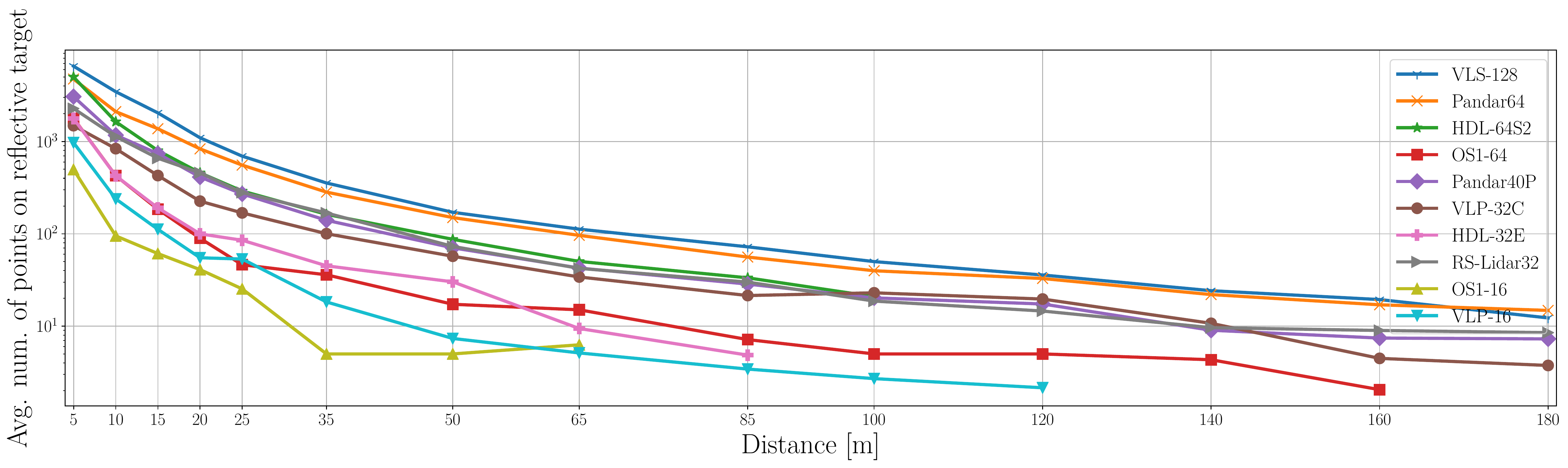}
		\label{F:density-b}
	}
	\caption{Expected vs measured density (number of points) on reflective targets per LiDAR, \subref{F:density-a} number of expected points, and \subref{F:density-b} average number of measured points.}
	\label{F:density}
	\vspace{-1.5em}
\end{figure*}

We used two metrics to compare LiDARs measurement performance: range accuracy and point density. We segmented the reflective targets as a whole and individually. We accumulated 40 frames of LiDAR data and rejected data with insufficient points (min. 9 points per target, or 3 points per reflective surface type). RMSE between the measured points and the ground truth was calculated at every distance, and results are shown in Fig.~\ref{F:range}. 
We can easily see that generally, RMSE grows with distance and some LiDARs struggle at very close distances. Upon closer investigation, some LiDARs specifically struggle with high reflectivity targets at close range. 

Fig.~\ref{F:density} shows the expected vs actual number of points detected on the reflective targets, averaged over 40 frames. The expected density is obtained from simulation, using each LiDAR's HRes, VRes and VFOV, to find the number of points falling inside the reflector targets at each range. In general, the VLS-128 had the best performance and measured values matched very closely the expected density. Pandar64 came in second place and also performed similar to its expected density. Pandar40P, RS-Lidar32 and VLP-32 followed closely the HDL-64S2. OS1-64 drops very rapidly within the first 20\,m and after 35\,m provides a similar density to the sensors with a lower number of channels. Finally, the OS1-16 comparatively had lower density than VLP-16 at the same number of channels. 

As shown on the VFOV data on Table~\ref{tab:lidar-list}, LiDARs have their laser vertical layout typically designed to cover more of the ground than the sky (i.e., more laser beams pointing downwards at negative elevation angles, than beams pointing upwards), the exception being VLP-16, OS1-16 and OS1-64 with symmetric coverage. Having the A0 reflective targets from 0.6\,m up to 1.8\,m from the ground while LiDARs are mounted on the car slightly over 2\,m from the ground, means that sensors which favour the ground portion will detect more points from targets than those which favour symmetric coverage. HDL-64S2 has the smallest sky portion coverage ($2\degree$) while HDL-32E has the largest ground portion coverage ($30.67\degree$).
In addition, VLS-128, VLP-32C, Pandar64, Pandar40P and RS-Lidar32 have the same VFOV of $40\degree$ with equal \vfovboundsdegplus{15}{25} bounds (of course, number of beams and VRes are different). These sensors have a rather high density performance up to the maximum range therefore are suitable for object detection. A detailed study of the vertical layouts and vertical resolutions of these LiDARs, for diverse applications such as object detection, object classification, mapping, localization, etc., is left as a future work.

\begin{figure*}[!htb]
	\centering
	\subfloat[][VLS-128]{
		\begin{minipage}[b]{.16\textwidth}
			\includegraphics[width=\textwidth]{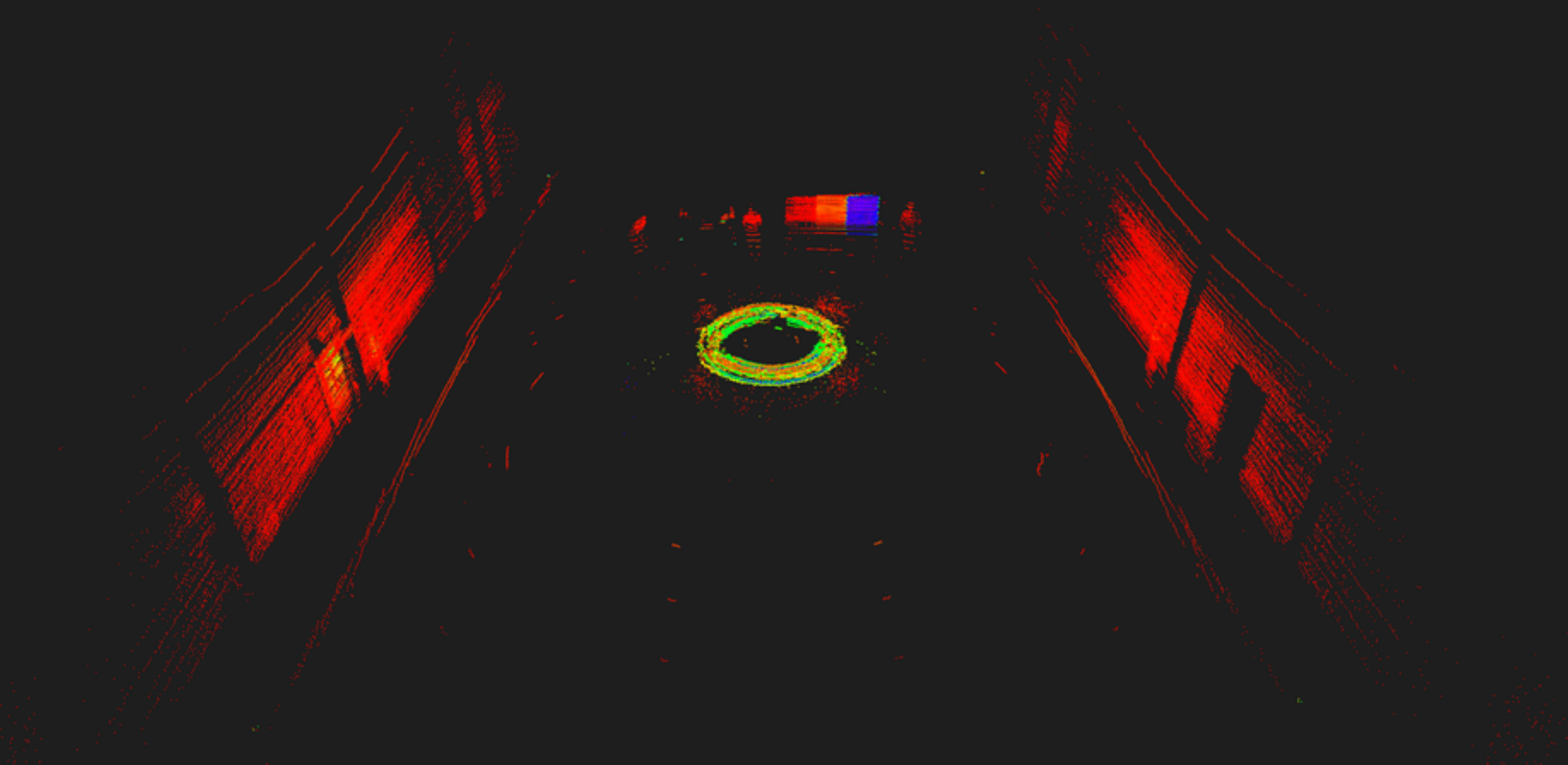}
			\vfill
			\includegraphics[width=\textwidth]{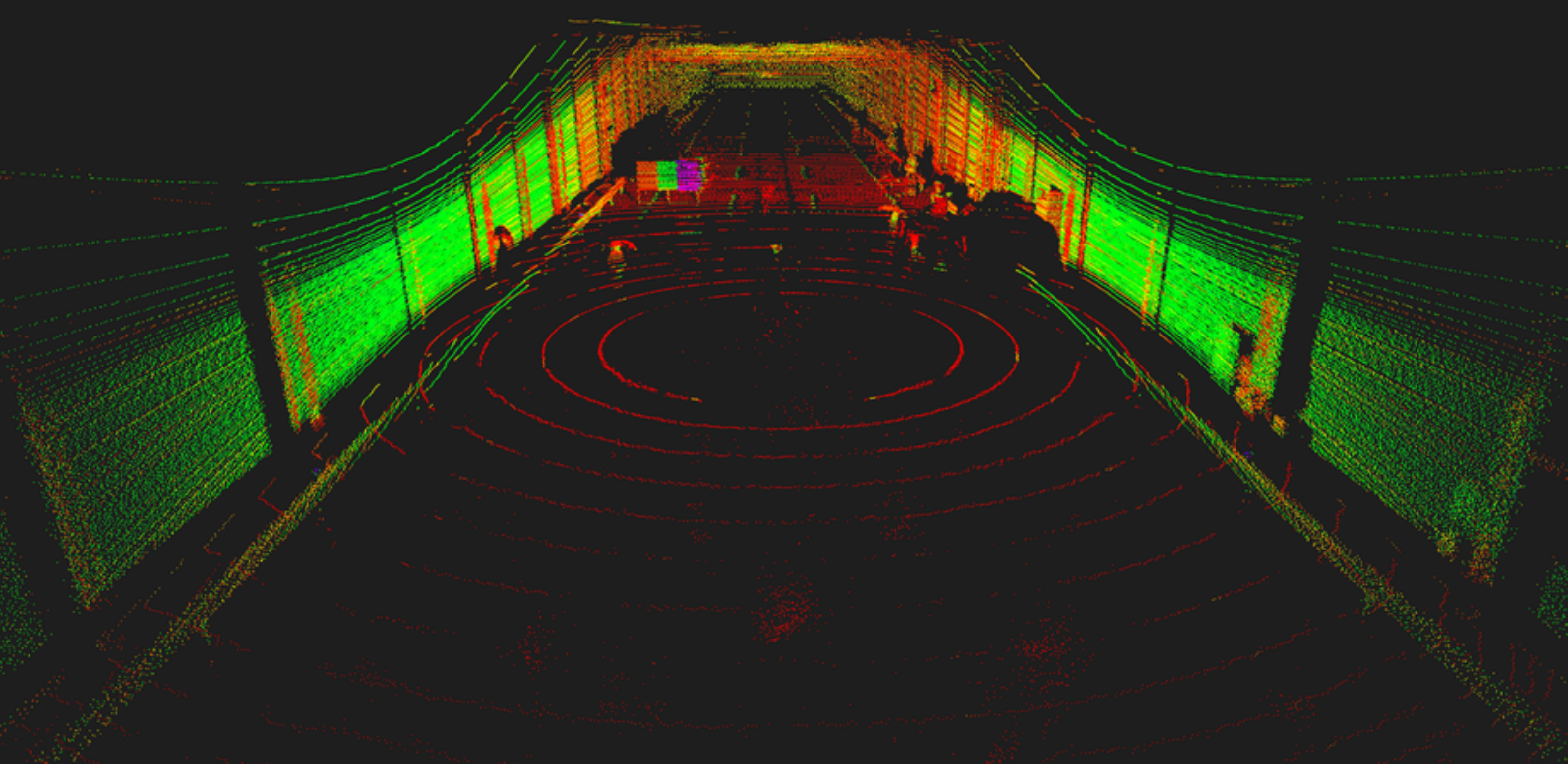}
			\vfill
			\includegraphics[width=\textwidth]{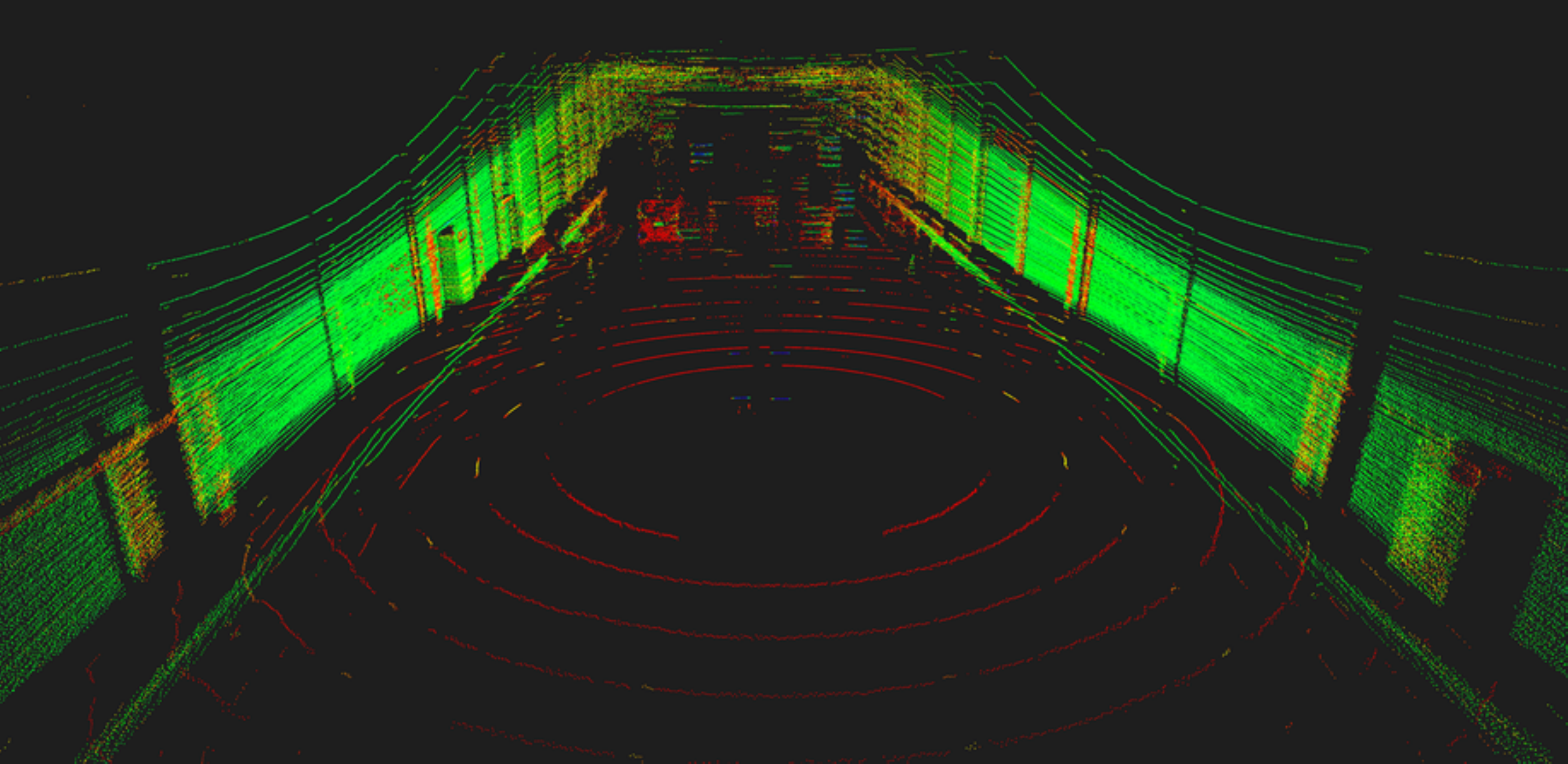}
		\end{minipage}
		\label{F:adverseweather-a}
	}\hspace{-1em}
	\subfloat[][HDL-64S2]{
		\begin{minipage}[b]{.16\textwidth}
			\includegraphics[width=\textwidth]{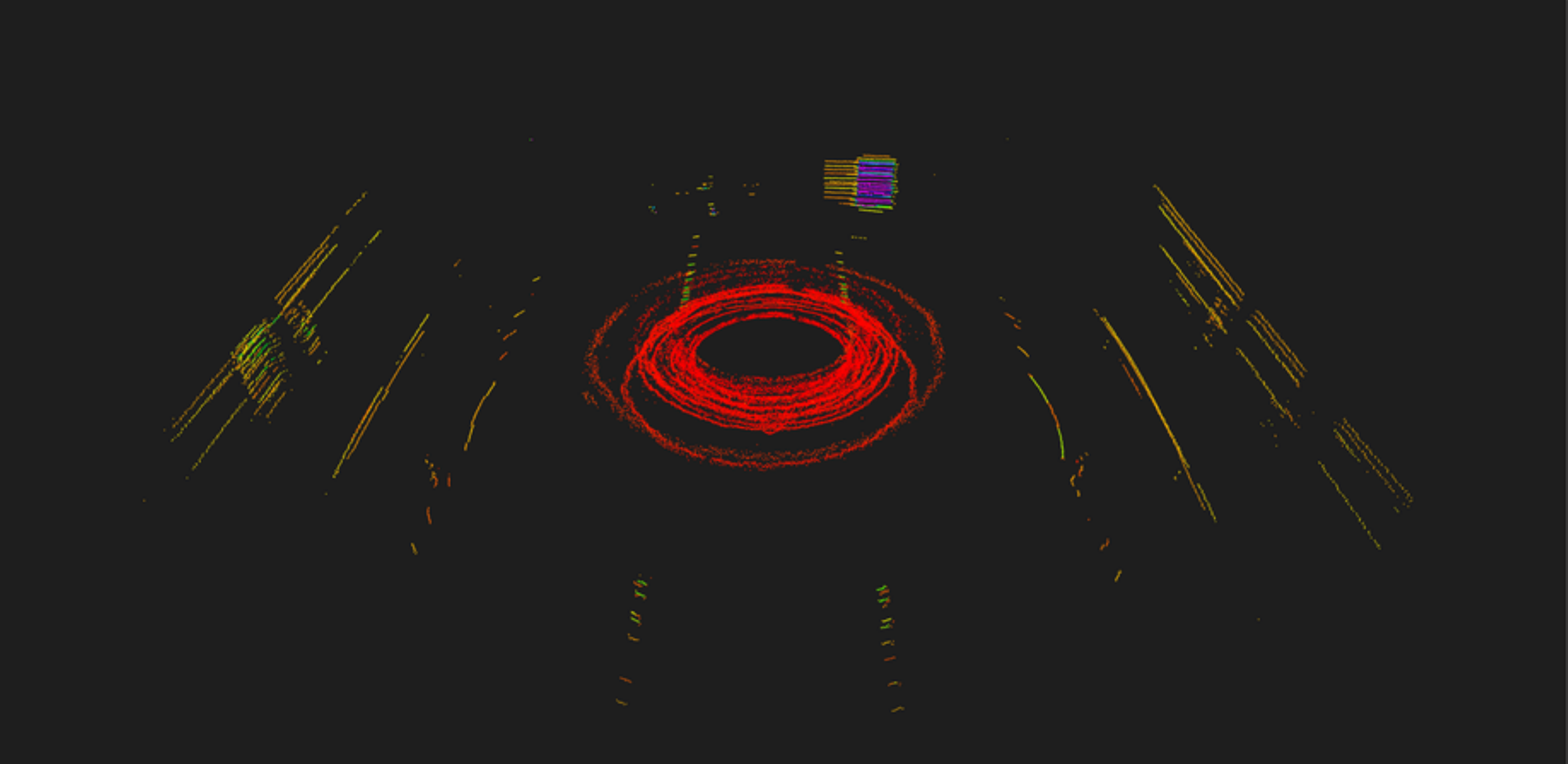}
			\vfill
			\includegraphics[width=\textwidth]{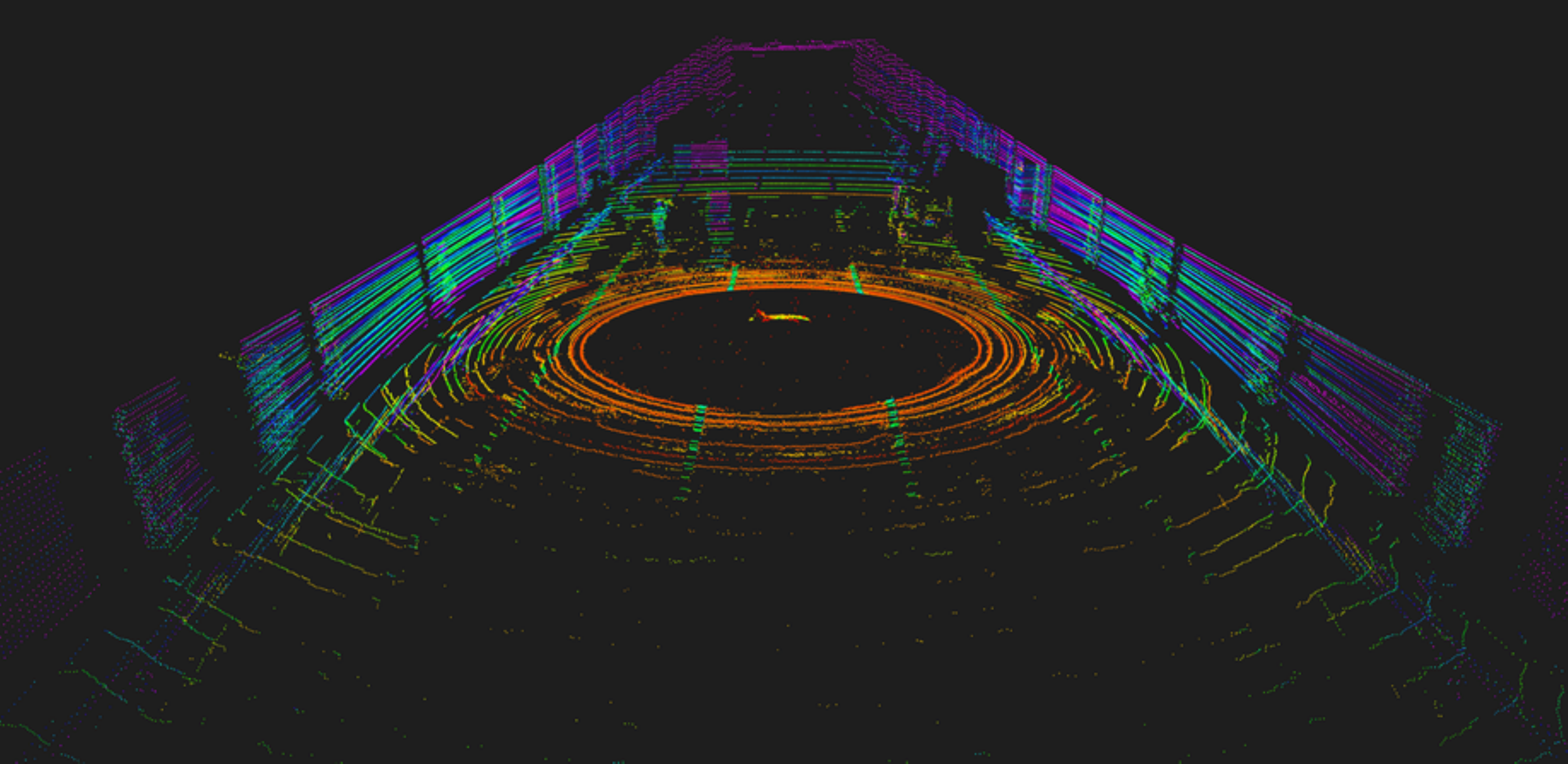}
			\vfill
			\includegraphics[width=\textwidth]{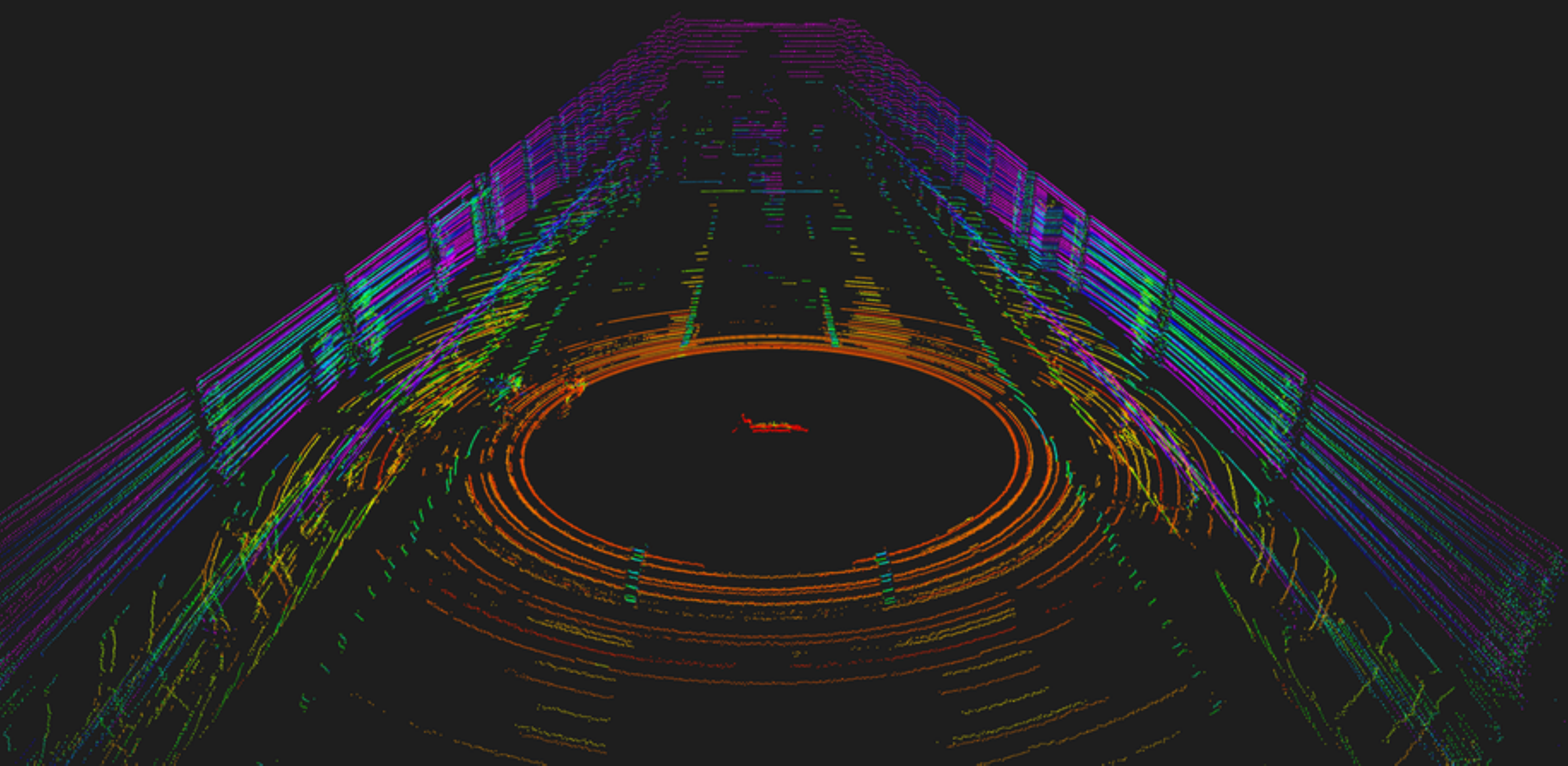}
		\end{minipage}
		\label{F:adverseweather-b}
	}\hspace{-1em}
	\subfloat[][HDL-32E]{
		\begin{minipage}[b]{.16\textwidth}
			\includegraphics[width=\textwidth]{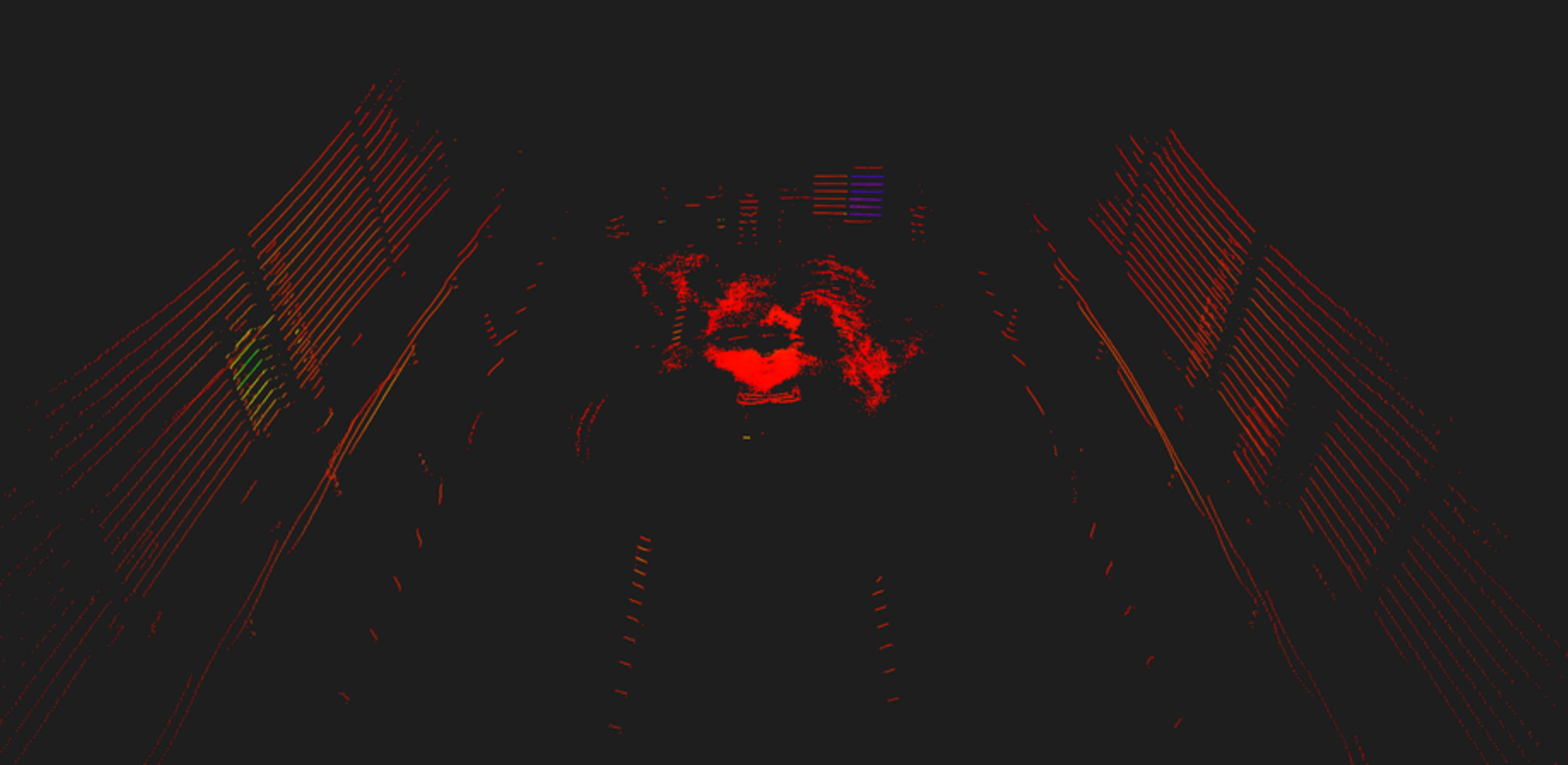}
			\vfill
			\includegraphics[width=\textwidth]{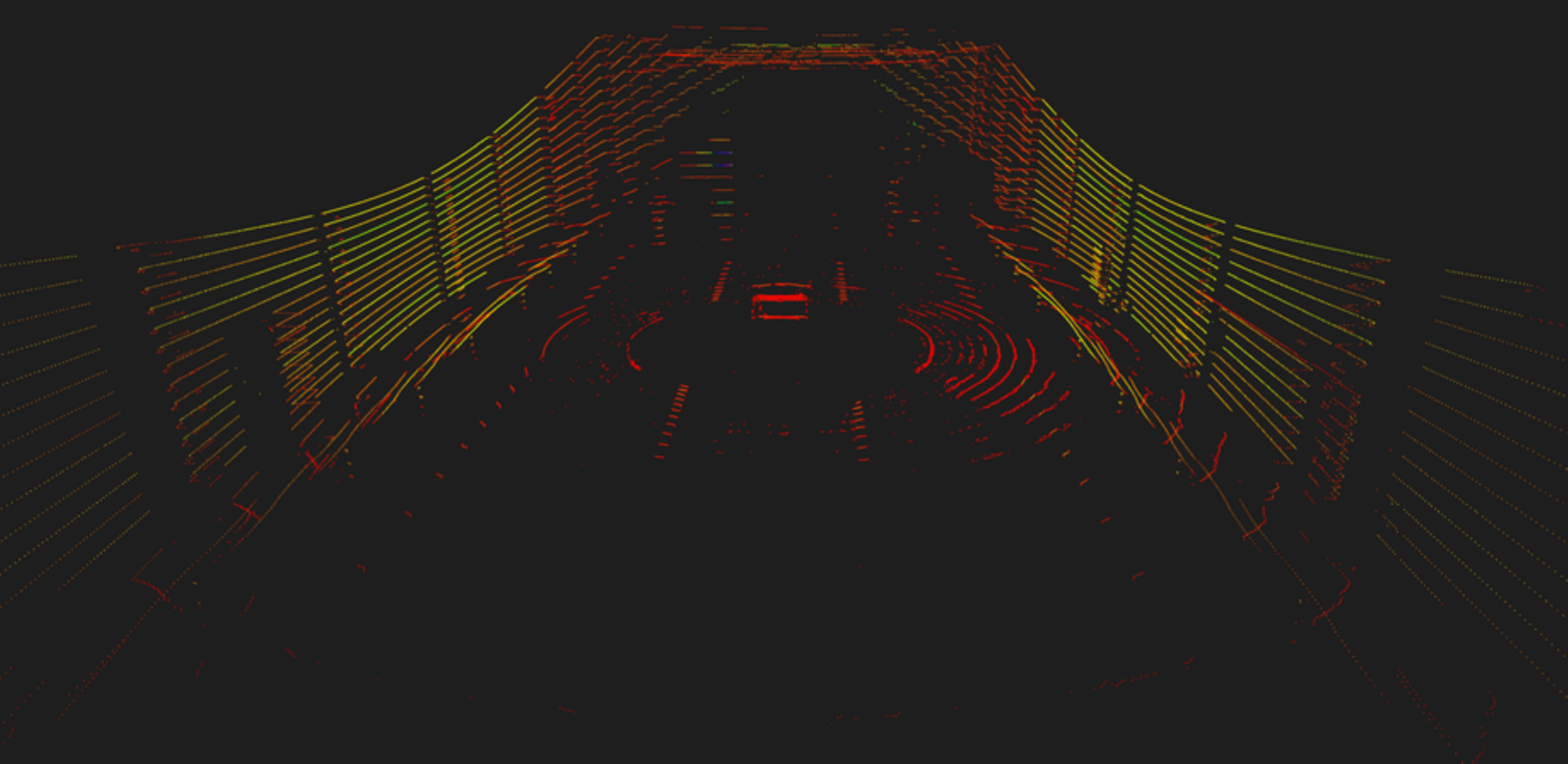}
			\vfill
			\includegraphics[width=\textwidth]{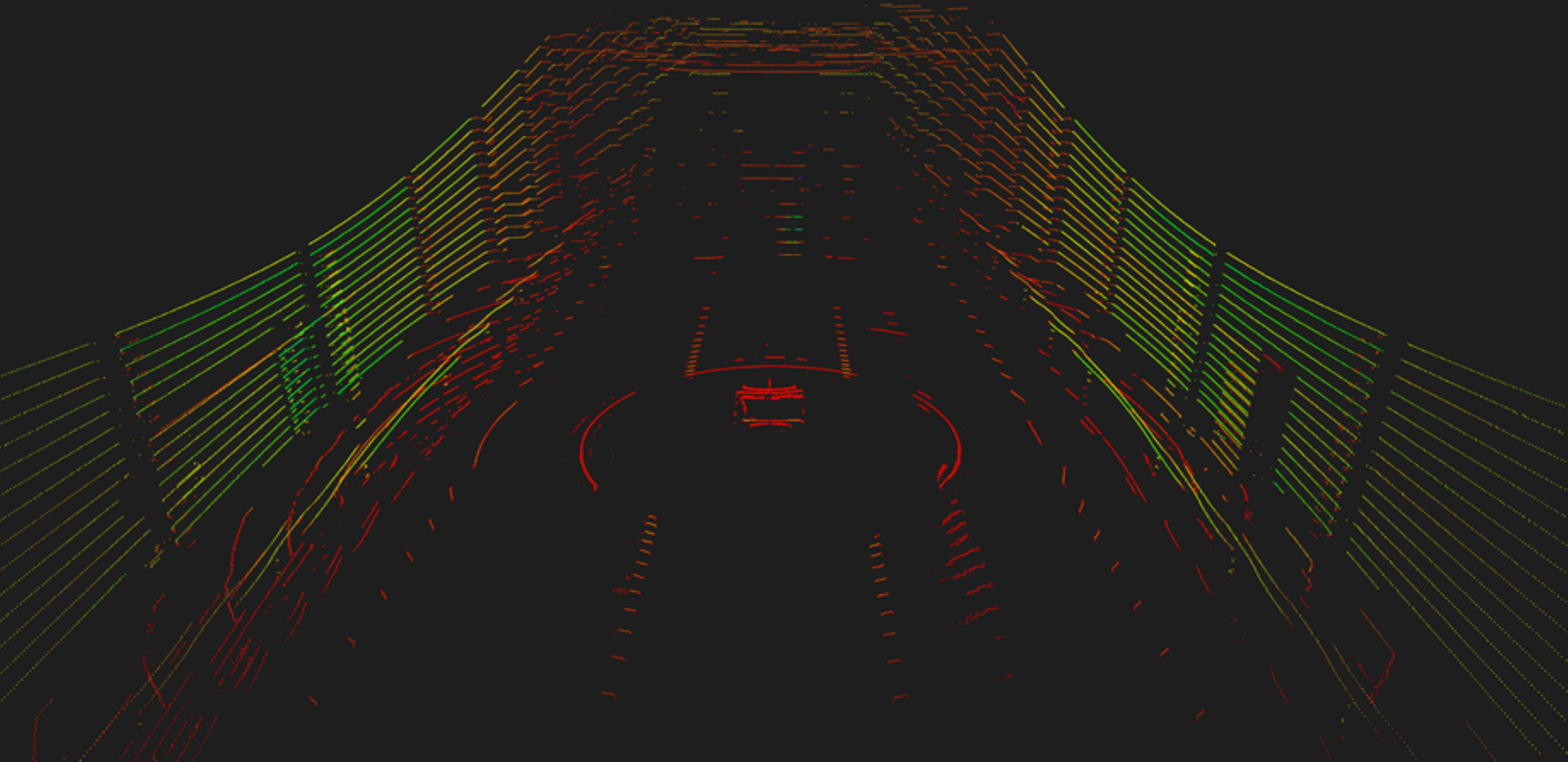}
		\end{minipage}
		\label{F:adverseweather-c}
	}\hspace{-1em}
	\subfloat[][Pandar64]{
		\begin{minipage}[b]{.16\textwidth}
			\includegraphics[width=\textwidth]{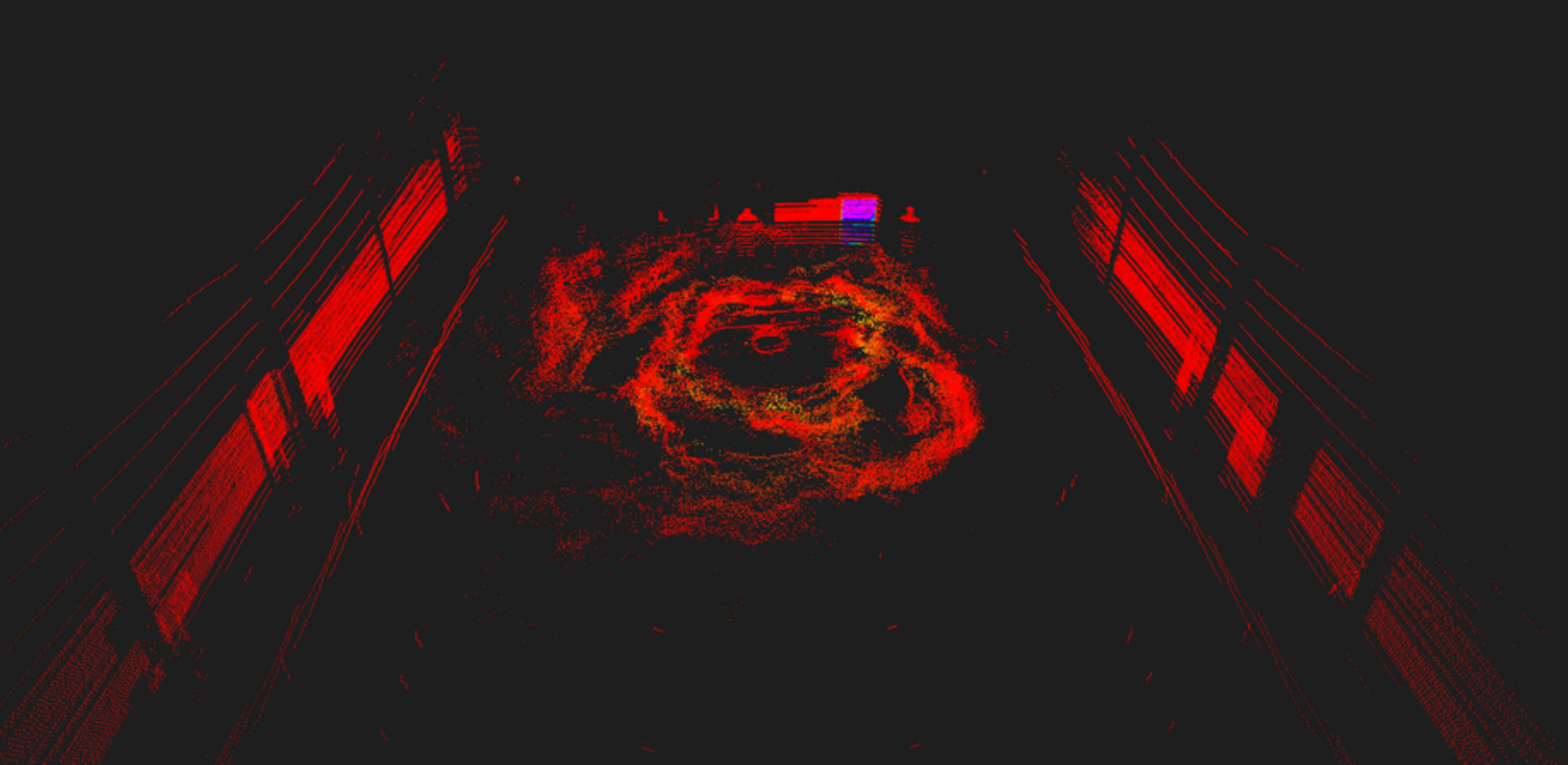}
			\vfill
			\includegraphics[width=\textwidth]{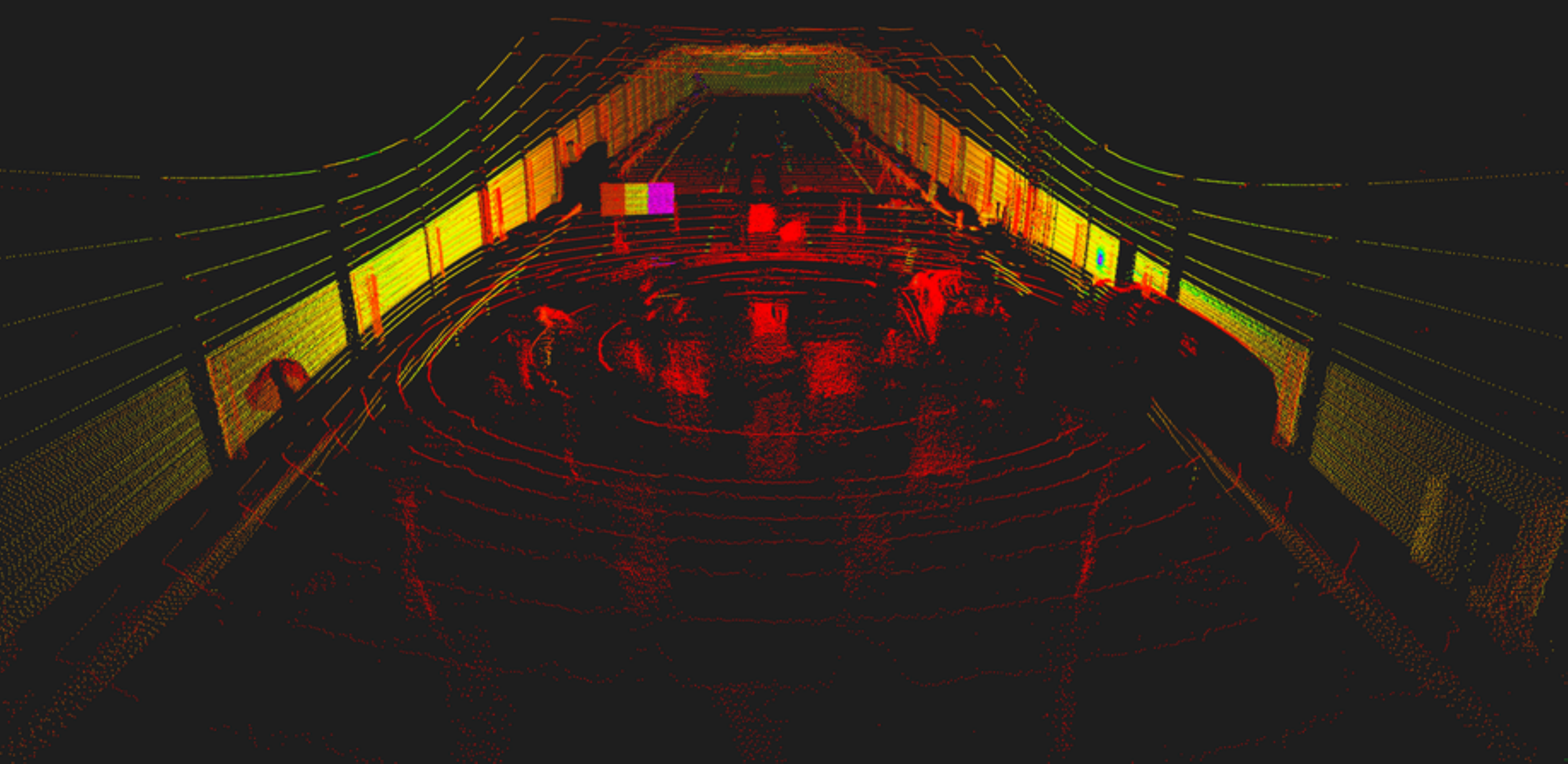}
			\vfill
			\includegraphics[width=\textwidth]{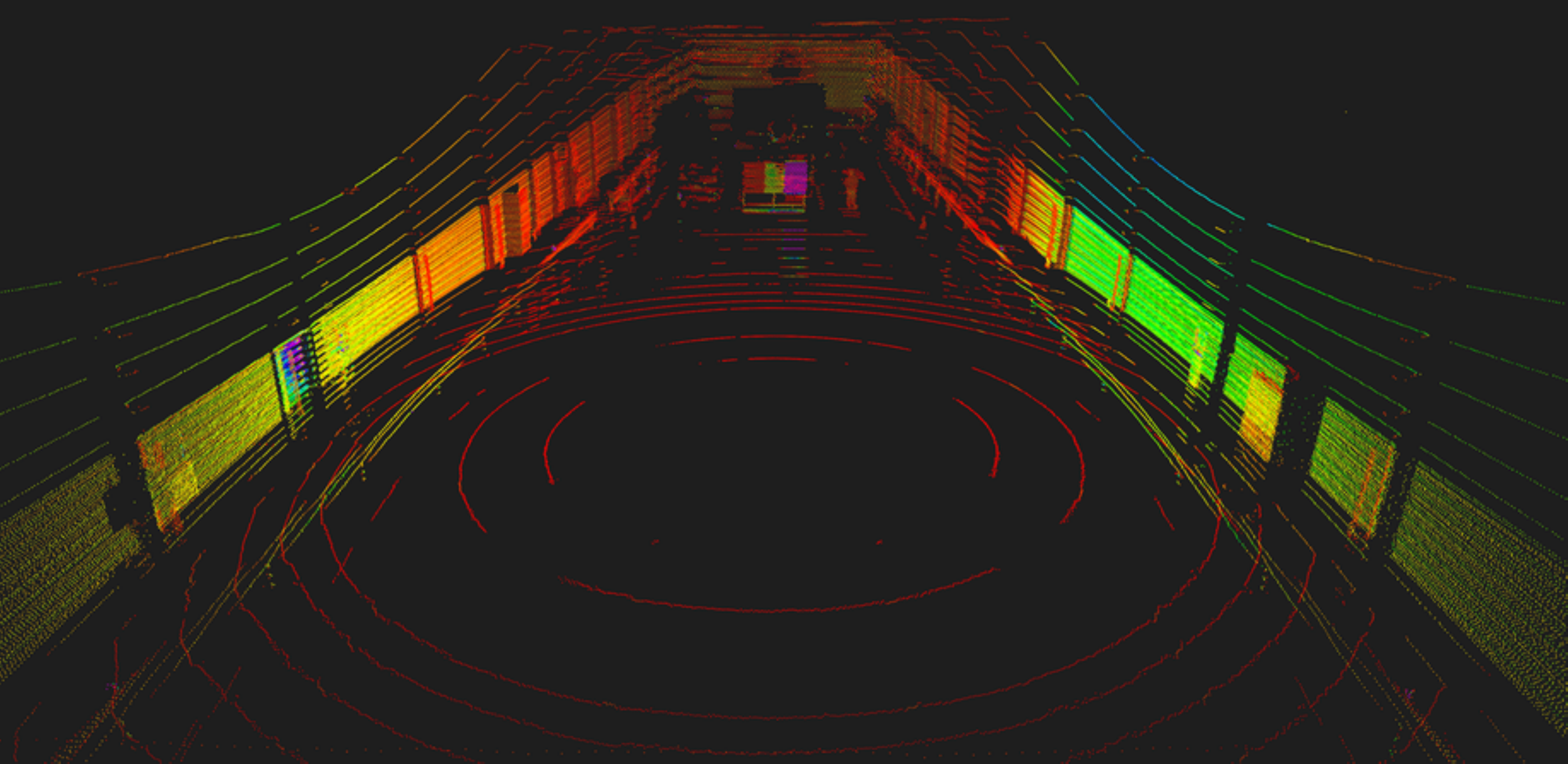}
		\end{minipage}
		\label{F:adverseweather-d}
	}\hspace{-1em}
	\subfloat[][RS-Lidar32]{
		\begin{minipage}[b]{.16\textwidth}
			\includegraphics[width=\textwidth]{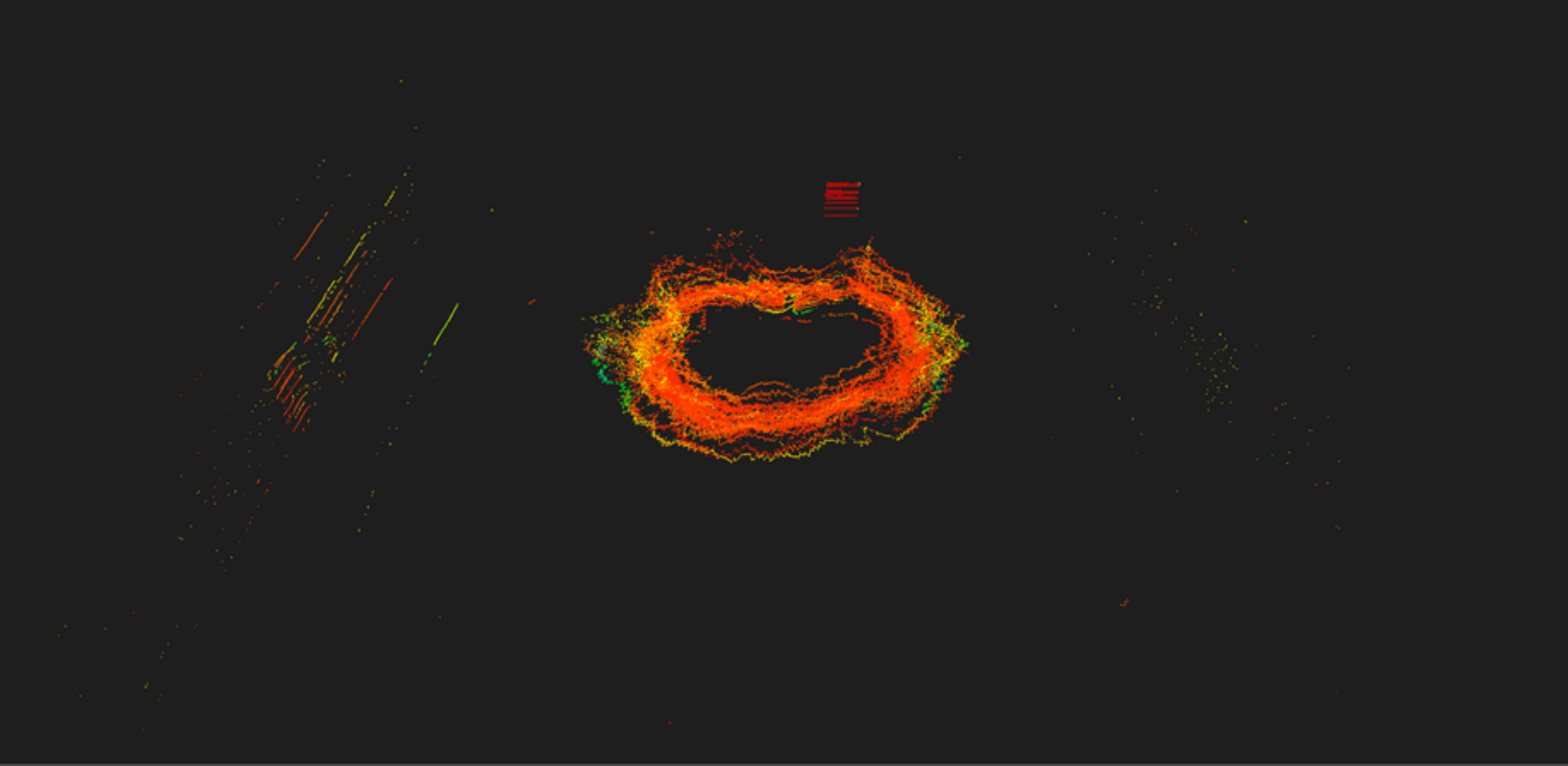}
			\vfill
			\includegraphics[width=\textwidth]{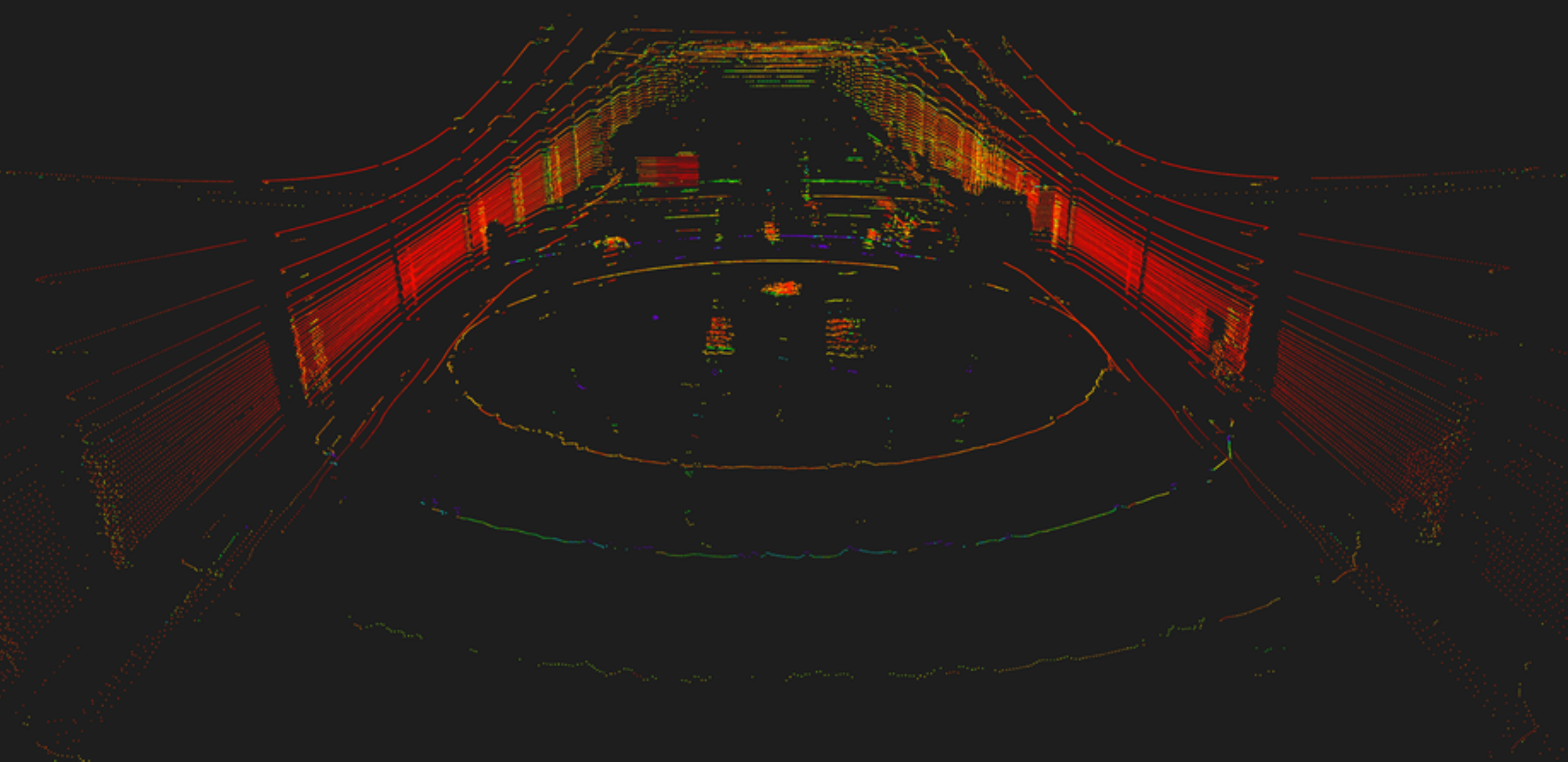}
			\vfill
			\includegraphics[width=\textwidth]{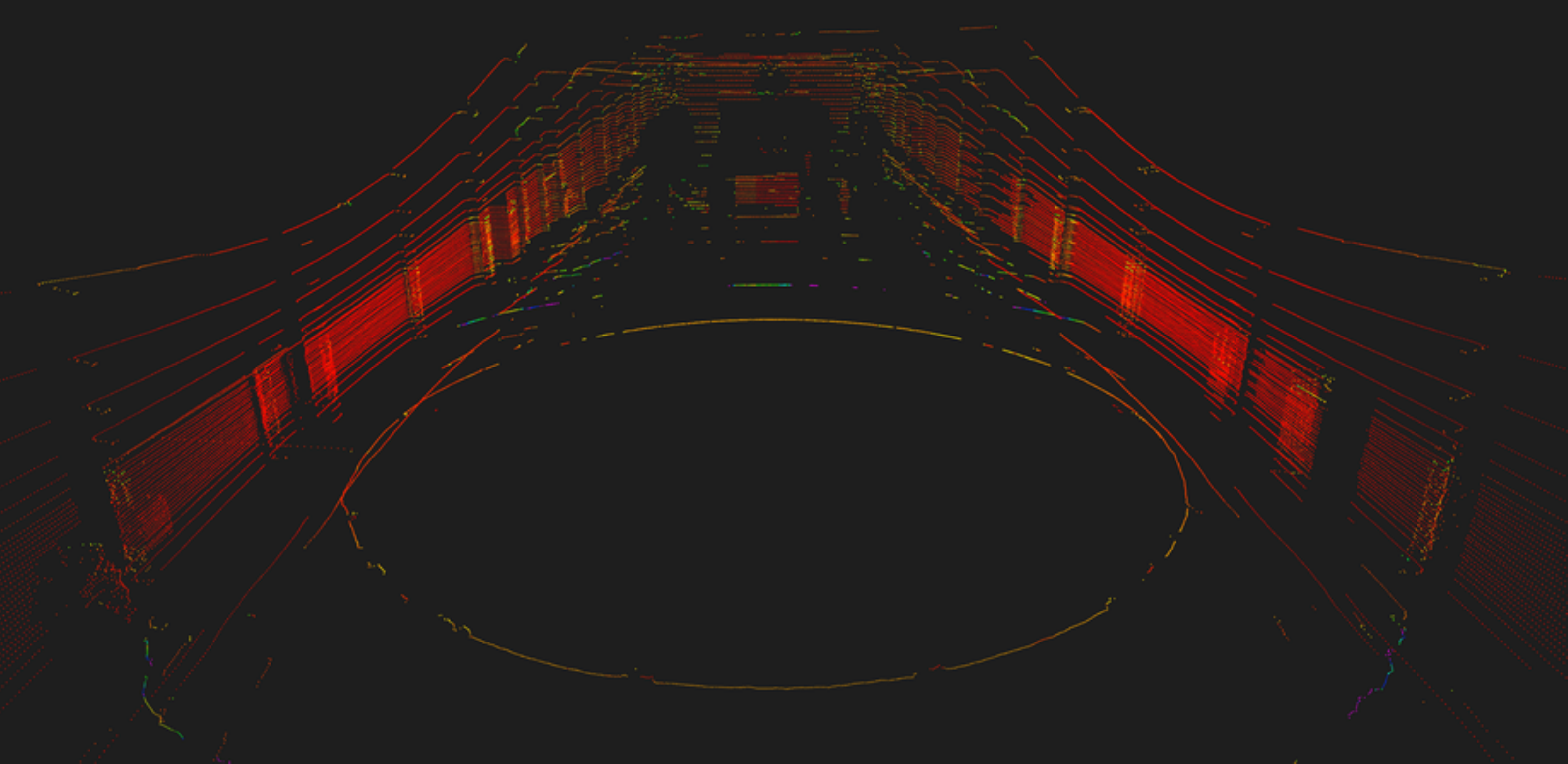}
		\end{minipage}
		\label{F:adverseweather-e}
	}\hspace{-1em}
	\subfloat[][OS1-64]{
		\begin{minipage}[b]{.16\textwidth}
			\includegraphics[width=\textwidth]{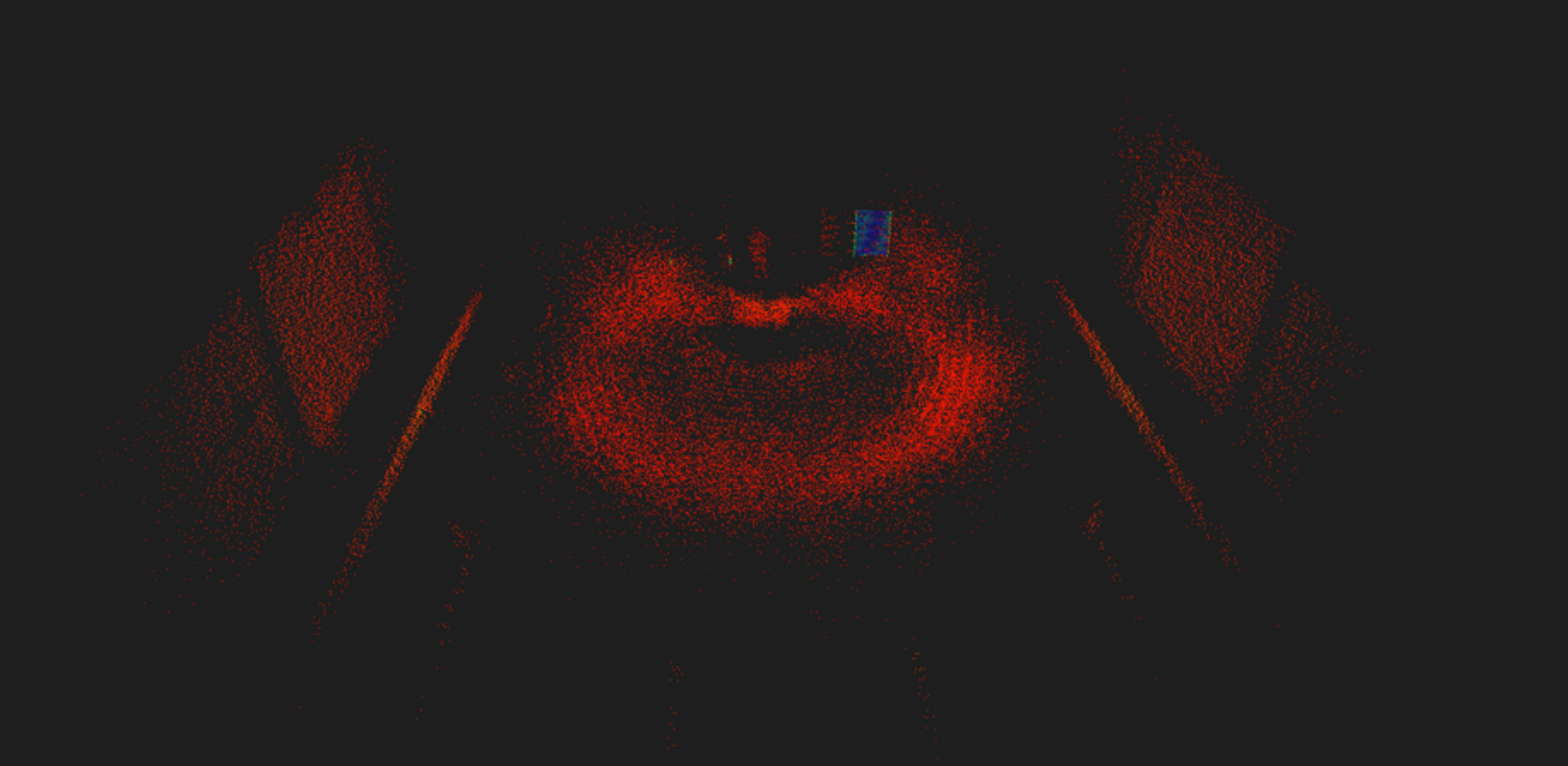}
			\vfill
			\includegraphics[width=\textwidth]{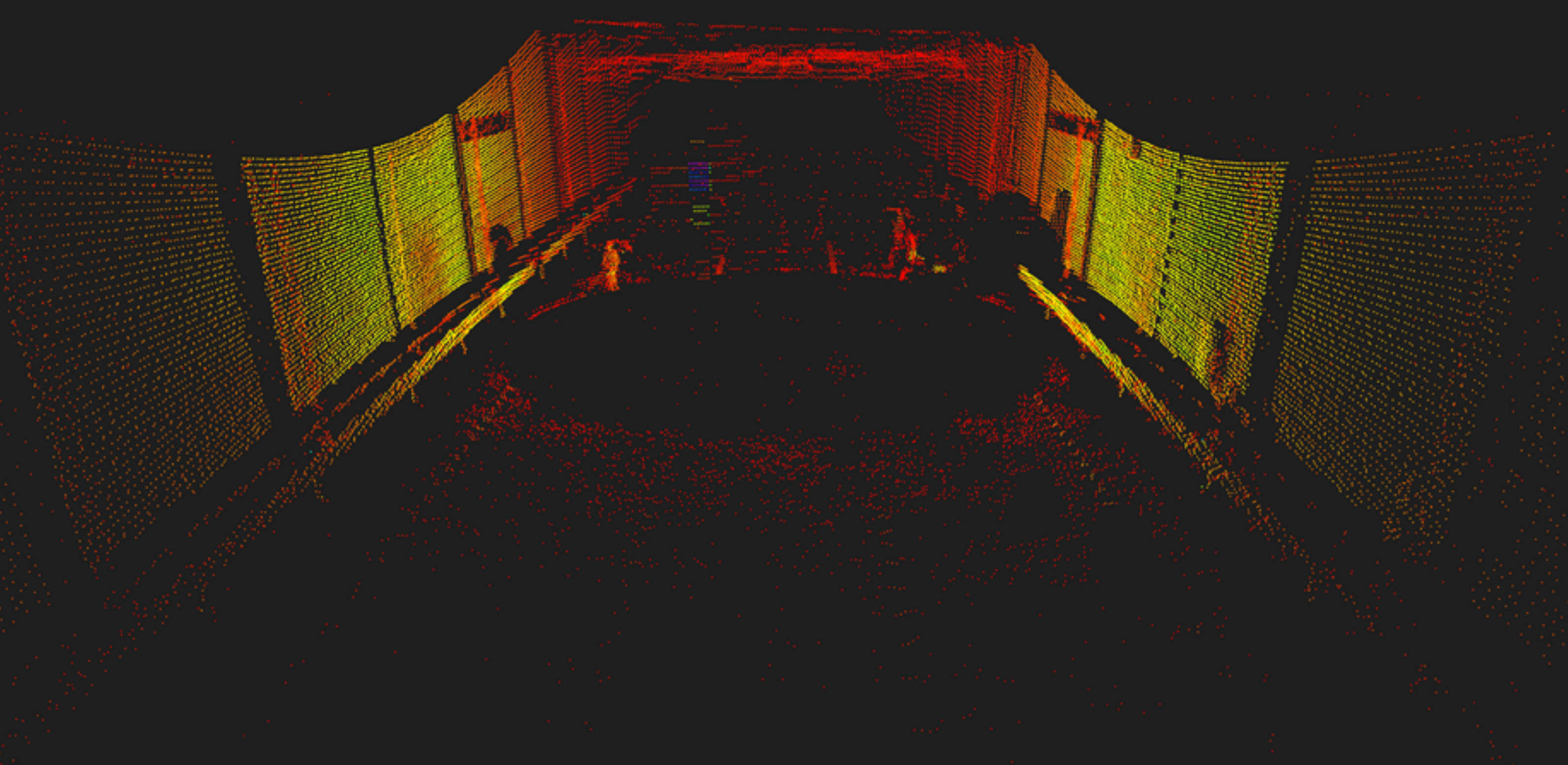}
			\vfill
			\includegraphics[width=\textwidth]{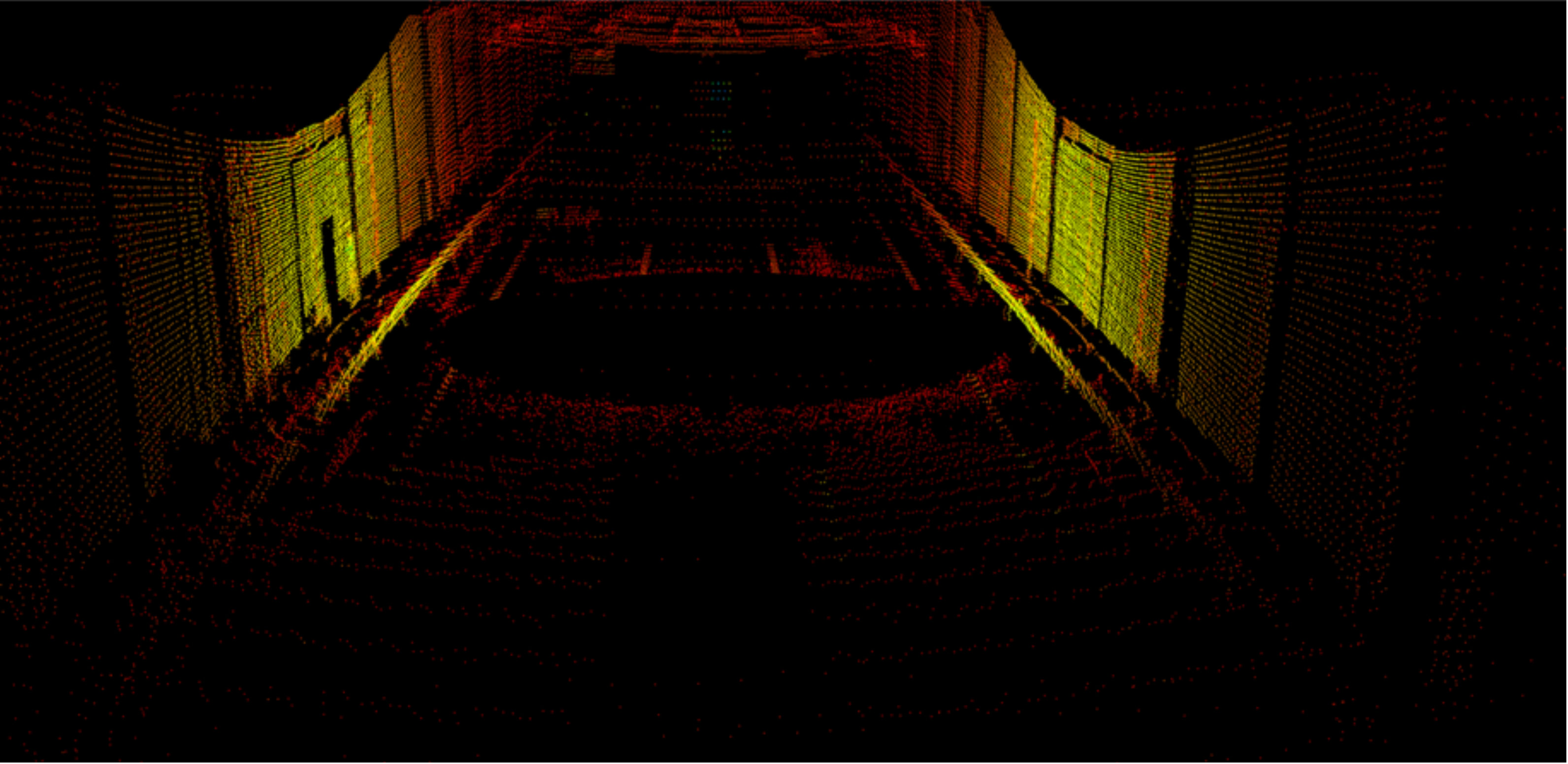}
		\end{minipage}
		\label{F:adverseweather-f}
	}
	\caption[]%
	{Adverse weather results, color represents intensity, top row: fog, middle row: rain, and bottom row: strong light.}
	\label{F:adverseweather}
	\vspace{-1.5em}
\end{figure*}
\section{Adverse weather}
\label{s:env-weather}
JARI's weather experimental facilities allowed us to test LiDARs in controlled weather conditions (refer to Fig.~\ref{F:jaritargets}\subref{F:jaritargets-a}). For fog emission, this weather chamber has 7.5\,$\mu$m particle size and controllable visibility of 10\,m up to 80\,m, with fog emitted over the complete 200\,m track. For rain emission, there are two different sprinklers with particle size of 640\,$\mu$m and 1400\,$\mu$m, and 3 precipitation levels: strong (30\,mm/h), intense (50\,mm/h), and very intense (80\,mm/h). In our study we used strong and very intense. Rain is emitted only for half of the track (100\,m). Strong ``sun'' light comes from a controlled mobile 6\,kW xenon light source with maximum luminous intensity of 350\,Mcd, and adjustable position, yaw and pitch angles. It has an optional yellow filter to approximate the color temperature of the sun; however, as it reduces illuminance, we tested without this filter for a maximum color temperature of 6000\,K (sample scene in Fig.~\ref{F:jaritargets}\subref{F:jaritargets-f}). In our experiment, the maximum illuminance at the LiDAR mount position on the car was set to 200\,klx (full sunlight illuminance at noon) at a distance of 40\,m from the light source. This means illuminance gradually increases from the starting position, reaching the peak illuminance at 40\,m from the light source, and then decreases towards the stopping position.

For safety reasons, during the adverse weather experiments, we drove the vehicle between 15\,km/h and 25\,km/h. Due to poor visibility during fog and light experiments, we also added small bumps on the road (see Fig.~\ref{F:jaritargets}\subref{F:jaritargets-d}) so the driver could identify the slow down and stopping positions; as we drove forward and backwards, there were two such stopping areas at either end of the track. For all the weather experiments, a passenger was present to lend an extra pair of eyes to the driver. The driver, other team members and the JARI staff kept constant communication over push-talk ratios to regulate the start and end of each run, and to ensure safety. For the fog experiment, we ensured fog density before each experiment. For the strong light experiment, both driver and passenger and other people outside the vehicle wore special dark sunglasses. The strong light experiment was conducted right after the rain experiment, thus our data has the additional value of including specular reflections (Fig.~\ref{F:jaritargets}\subref{F:jaritargets-e}) due to the wet road surface for half the test track. We also recorded RGB camera and IR camera data during these experiments.

The fog experiment started with a very dense 10\,m visibility and the vehicle drove forward towards the stop position, then backwards towards the start position, waited 30\,s for the fog to dissipate, and repeated again until perceived visibility was over 80\,m. It takes about 10\,min for the fog chamber to reach maximum density again, so during this time we changed LiDARs (we kept other LiDARs warming up at least 30\,min before any test) and repeated. For the rain experiment, we started with a 30\,mm/h precipitation rate, waited about 1\,min for it to become steady, and drove the vehicle backwards towards the stop position and then forward to the start position only one time; as rain falls only in the last half of the track, our vehicle made transitions from dry to rainy and vice versa, with targets inside the rainy area. We then set the 80\,mm/h precipitation rate and repeated driving, returning to the start position to change LiDARs for the next test. Finally, the strong light experiment happened after rain experiment therefore half the test track was wet creating specular reflection conditions; from the start position we turned on the xenon light source, drive forward towards the stop position (passing through the maximum illuminance zone) and backwards towards the start position, turned off the light, changed the LiDAR, and repeated.

Adverse weather qualitative results are shown on Fig.~\ref{F:adverseweather} for a selection of LiDARs. The top row shows the fog experiment when the vehicle was close to the targets, the middle row shows the rain experiment at 30\,mm/h precipitation rate with the vehicle under the rainy area, and the bottom row shows the strong light experiment when the vehicle was close to the highest illuminance area. All LiDARs were affected in a similar way by fog: several of the low reflection intensity points tend to form a toroidal shape around the LiDAR for the echo from the fog is stronger, the highly reflective walls are partially visible but with a much lower intensity values, with only the highly reflective white markers in the road and the diamond-grade and white reflectors ahead are partially visible with a diminished intensity; this means that much of the intensity of the reflected light is scattered and attenuated by the fog.
\begin{figure}[!htb]
	\centering
	\includegraphics[width=0.4\textwidth]{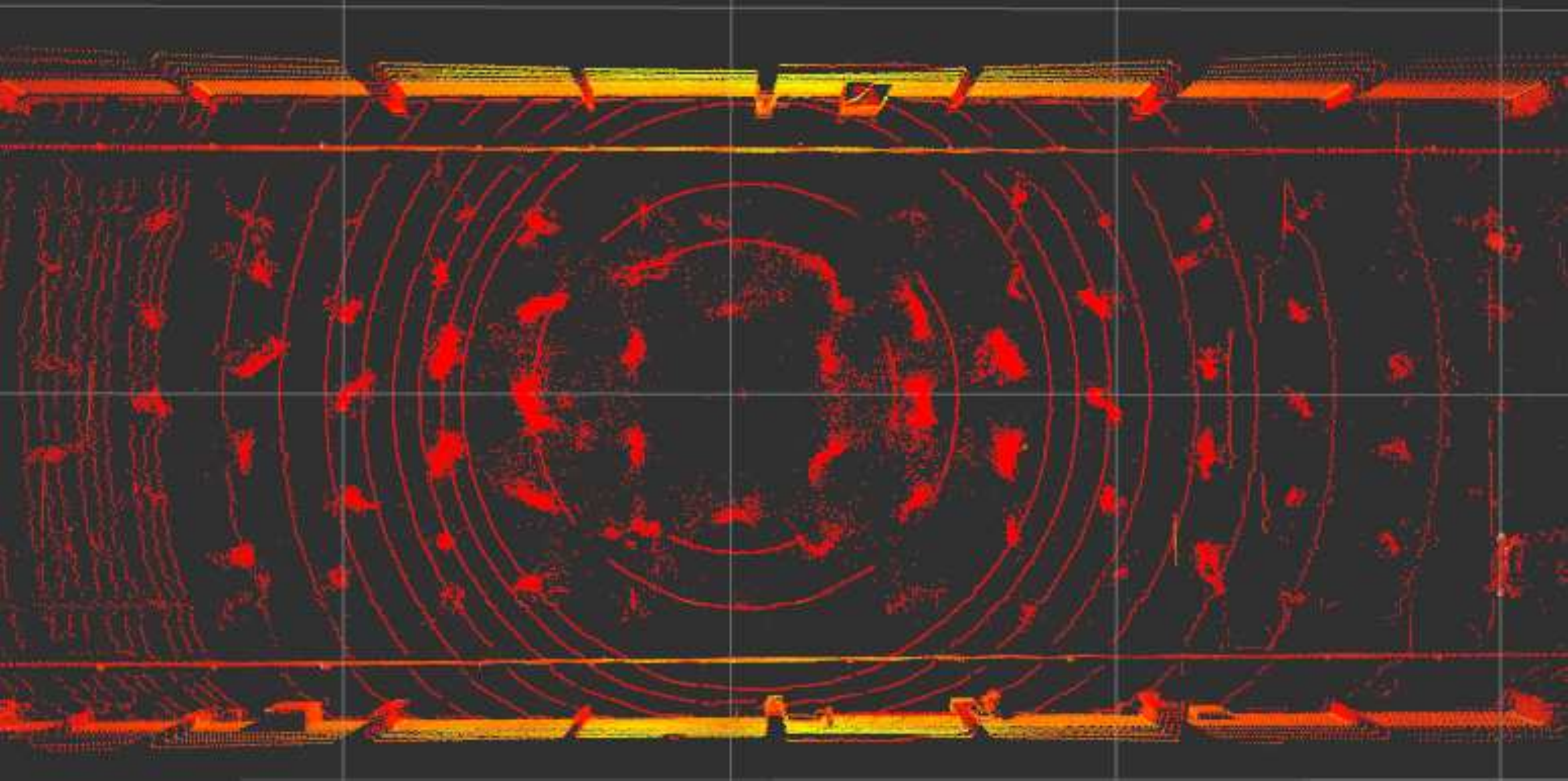}
	\caption{``Rain pillars'' as detected by a LiDAR.}
	\label{F:rain}
	\vspace{-1.5em}
\end{figure}
Rain also affects all the LiDARs: while it does not attenuate reflections, it creates fake obstacles especially when precipitation rate is high and non-uniform. This situation is clearly shown in Figs.~\ref{F:adverseweather}\subref{F:adverseweather-a}, \subref{F:adverseweather-d} and \subref{F:adverseweather-e}. The rain experiment was not encouraging, as most LiDARs detected the water showers from the sprinklers as vertical pillars, as shown in Fig.~\ref{F:rain}. This points to the need of better rain generation systems in weather chambers.

Finally, during the strong light experiment, when the vehicle was approximately at the maximum illuminance area, we obtained almost no data from the experiment targets, road and wall in front of the LiDAR. These elements become again visible when the vehicle is in other areas with much lower illuminance. While such strong illuminance is not expected at the horizon, certain LiDAR setups on the car, especially when LiDARs are mounted with large roll/pitch angles, will be affected by strong sunlight.

\section{Conclusions}
\label{s:concl}
In this work we introduced an first-of-its-kind collection of data from multiple 3D LiDARs, made publicly available for research and industry, with the objective of improving our understanding of the capabilities and limitations of popular LiDARs for autonomous vehicles. This dataset will enable benchmarking of new LiDARs, better representations in vehicle simulation software, direct comparison of LiDARs capabilities before purchasing, and better perception algorithms. 

This study still lacks some important conditions such as low temperature snowy environments, night time scenes, direct interference, realistic rain, other wavelengths, and so on, which will be addressed in future extensions. However, this work sheds light onto existing issues with LiDARs which require research: the serious noise induced by indirect interference and strong light, the almost null visibility during dense fog, and the need for adequate existing object detection algorithms to work with multiple LiDARs. 

Efforts to extend the {\datasetname} dataset have already started: we will add more sensors, including the Velodyne Alpha Prime; environments, including direct and indirect interference; other evaluations, including new perception algorithms, mapping and localization. We are also preparing a second phase which will include, among others, newer solid-state LiDARs (MEMS-based scanners), different wavelengths such as 1550\,nm, and other scanning techniques.

\section*{Acknowledgments}
This work was supported by the Core Research for Evolutional Science and Technology (CREST) project of the Japan Science and Technology Agency (JST). 
We would like to extend our gratitude to the Japan Automobile Research Institute (JARI) for all the support while using their JTOWN weather chamber and other facilities. We appreciate the help of the Autoware Foundation 
to realize this project. Finally, and as a matter of course, this work would not have been possible without the invaluable support of Velodyne Lidar Inc., Ouster Inc., Hesai Photonics Technology Co., Ltd., and RoboSense--Suteng Innovation Technology Co., Ltd.

\bibliographystyle{IEEEtran}
\bibliography{IEEEabrv,multiple-3d-lidar}

\end{document}